\documentclass[11pt]{article}

\usepackage[final]{acl}

\usepackage{times}
\usepackage{latexsym}
\usepackage[T1]{fontenc}
\usepackage[utf8]{inputenc}
\usepackage{microtype}
\usepackage{inconsolata}
\usepackage{graphicx}
\usepackage{booktabs}
\usepackage{multirow}
\usepackage{amsmath,amssymb}
\usepackage{caption}
\usepackage{pifont}
\usepackage{enumitem}
\usepackage{makecell}
\usepackage[table]{xcolor}
\usepackage{pifont}
\usepackage{fvextra}
\newcommand{\myrothead}[2][60]{\rotatebox[origin=c]{#1}{\makecell[c]{#2}}}

\usepackage{tikz}
\usetikzlibrary{calc}

\usepackage{nicematrix} 
\NiceMatrixOptions{cell-space-limits=2pt} 

\title{FBS: Modeling Native Parallel Reading inside a Transformer}

\author{Tongxi Wang \\
  Southeast University / Nanjing, China \\
  \texttt{tongxi\_wang@seu.edu.cn} \\}

\begin{document}
\maketitle

\begin{figure*}[ht]
  \centering
  \includegraphics[width=\linewidth]{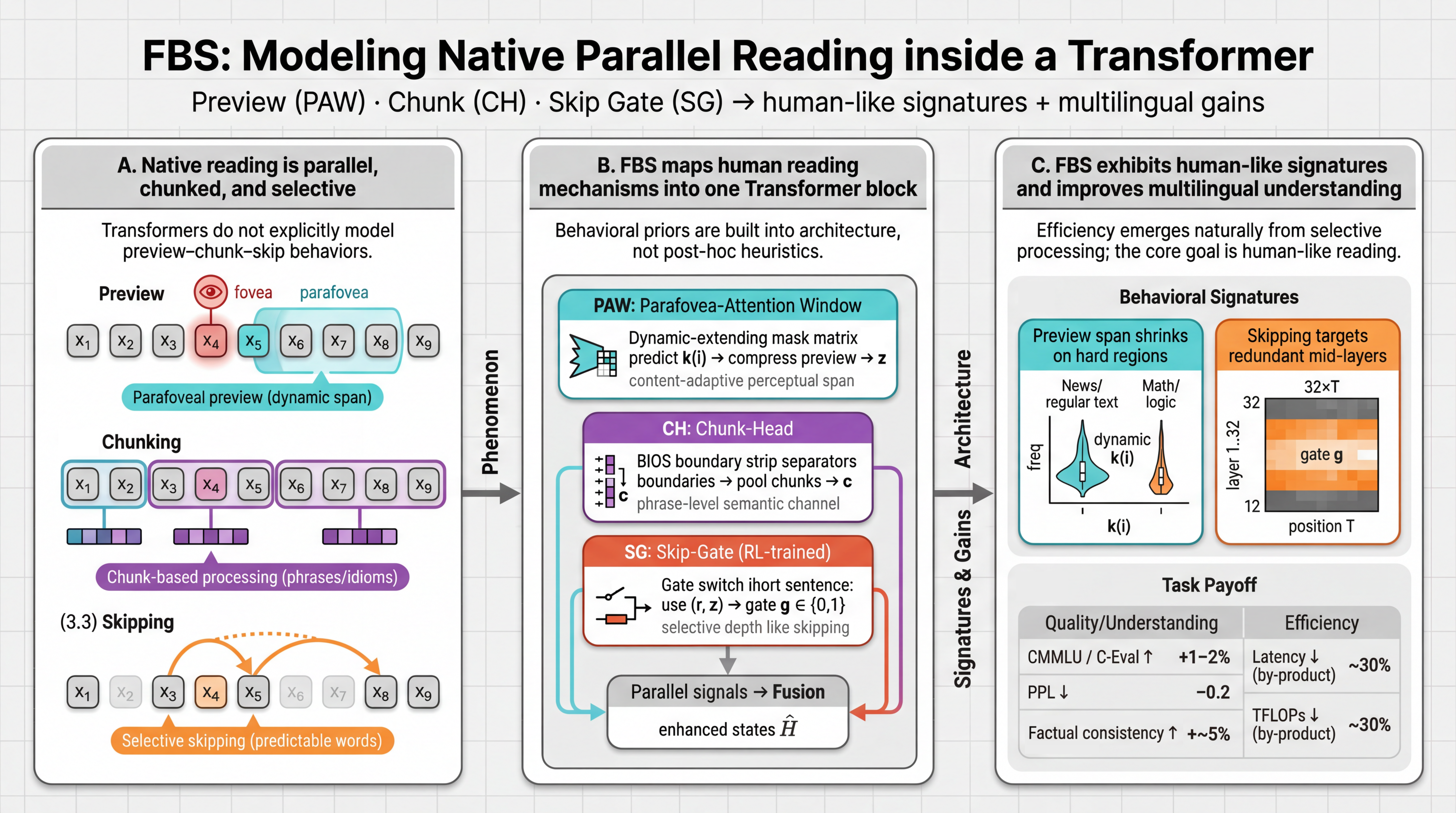}
  \caption{\textbf{FBS block overview.}}
  \vspace{-14pt}
\end{figure*}

\begin{abstract}
Large language models (LLMs) excel across many tasks, yet inference is still dominated by strictly token-by-token autoregression. Existing acceleration methods largely patch this pipeline and miss core human-reading ingredients: content-adaptive foresight, chunk-structure-aware compute allocation, and train-test consistency for preview/skimming. We propose the \textbf{Fovea-Block-Skip Transformer} (FBS), which injects a causal, trainable loop into Transformers via Parafovea-Attention Window (PAW), Chunk-Head (CH), and Skip-Gate (SG). Across diverse benchmarks, FBS improves the quality-efficiency trade-off without increasing parameters, and ablations show the three modules are complementary.
\end{abstract}

\section{Introduction}
\label{sec:intro}

Large language models (LLMs) have achieved strong performance in dialog, code generation, and cross-domain reasoning~\cite{Minaee2024LargeLMA,Touvron2023LLaMAOAA,guo2025deepseek}. Yet at inference time, most LLMs still advance through a strictly token-by-token autoregressive process: computation is spent almost uniformly, step after step, along a single causal chain. This stands in sharp contrast to proficient human reading, especially by natives, which is often \textbf{highly parallel} and \textbf{multi-granular}: readers leverage parafoveal preview~\cite{xu2022word,Touvron2023LLaMAOAA,snell2017evidence}, chunk-level processing~\cite{yang2020we,BazantKimmel2018LearningTRA}, and semantic-load-adaptive skimming~\cite{Fitzsimmons2014SkimRAA,Duggan2009TextSTA} to improve efficiency while maintaining comprehension quality.

A large body of ``faster decoding'' work improves efficiency by patching the autoregressive pipeline~\cite{chen2023accelerating,wang2025think}, e.g., draft-and-verify speculative decoding~\cite{leviathan2023fast,shen2026double}, multi-candidate generation (Medusa/EAGLE)~\cite{cai2024medusa,li2024eagle,li2024eagle2}, set/block decoding~\cite{gat2025set,zhang2025pdtrim,limol}, and adaptive computation such as early-exit or dynamic depth~\cite{xin2020deebert,teerapittayanon2016branchynet}. These methods are effective, but they typically miss three structural ingredients that make human reading both fast \emph{and} reliable: \textbf{(i) content-adaptive, structured foresight} (look-ahead is often fixed or external), \textbf{(ii) coupling between chunk structure and compute allocation} (chunks are used for representation compression rather than for deciding where to allocate compute), and \textbf{(iii) a matched train-test pathway} (previewing/skimming is often introduced only at inference, risking distribution shift).

We argue that breaking the autoregressive bottleneck is less about stacking isolated tricks and more about building an \emph{endogenous} mechanism that preserves causality while realizing a native-speaker-style pipeline: \textbf{preview $\rightarrow$ chunking $\rightarrow$ skimming}. Concretely, an efficient generator should (1) form a low-resolution \emph{predictive preview} to guide upcoming decisions, (2) integrate local semantics at the level of \emph{chunks} rather than only tokens, and (3) regulate compute \emph{dynamically}—skimming stable regions and pausing to compute when semantic load or uncertainty rises.

Guided by this viewpoint, we propose the \textbf{Fovea-Block-Skip Transformer (FBS)}, a causal Transformer augmented with a trainable, verifiable reading pipeline. FBS consists of three cooperative modules: \textbf{Parafovea-Attention Window (PAW)} produces \emph{content-adaptive predictive preview} from the model's own next-token distributions; \textbf{Chunk-Head (CH)} builds an online chunk-level semantic channel for phrase-scale integration; and \textbf{Skip-Gate (SG)} turns these signals into \emph{true layer/block skipping} during decoding, allocating computation where it matters. The three modules form a closed loop: preview informs chunking, chunk semantics stabilize decisions, and the resulting confidence/semantic load controls skimming.

We instantiate FBS by continual pre-training \texttt{Qwen3-4B-Instruct}~\cite{yang2025qwen3} on a Chinese–English mixed corpus and evaluate it on a broad set of reasoning, knowledge, math, and code benchmarks. Under a matched-parameter setting, FBS improves quality on major benchmarks while reducing executed compute; in a 512$\rightarrow$128 generation protocol, it achieves substantial wall-clock latency reduction (about 30\%) and significantly lowers TFLOPs. Analyses further show that PAW, CH, and SG are complementary: PAW provides foresight, CH strengthens local semantic organization, and SG contributes the dominant efficiency gains, yielding human-like ``speed up on stable spans, slow down at critical points'' behavior.

Our main contributions are:
(i)We formalize a \emph{native-speaker parallel reading} abstraction for LLM inference and translate it into trainable mechanisms suitable for causal Transformers.
(ii)We propose \textbf{FBS}, a unified architecture that couples PAW and CH with an internal compute controller SG, yielding an endogenous \textbf{preview$\rightarrow$chunk$\rightarrow$skim} loop that preserves causality while enabling real block/layer skipping.
(iii)We demonstrate consistent \textbf{quality-efficiency} gains on Chinese and English benchmarks under a matched-parameter setting and provide analyses that explain when and why the model skims or computes more.

\begin{figure*}[ht]
  \centering
  \includegraphics[width=\linewidth]{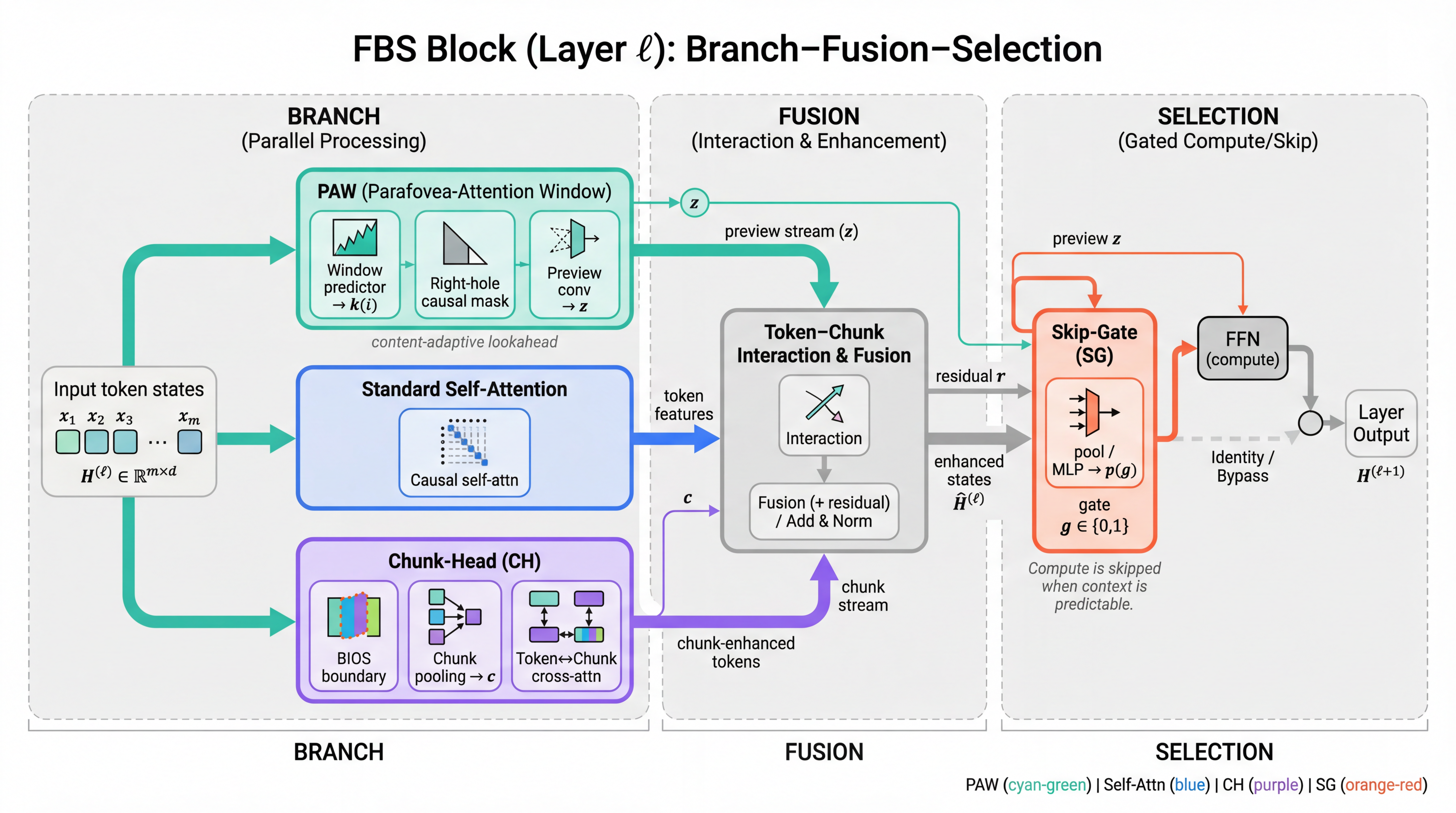}
  \caption{\textbf{FBS Pipeline}}
  \label{fig:fbs_block_overview}
  \vspace{-10pt}
\end{figure*}

\section{Methodology}
\label{sec:method}
We present the \textbf{Fovea-Block-Skip Transformer (FBS)}, which augments each causal Transformer layer with three lightweight modules: \textbf{PAW}, prefix-only predictive preview with a content-adaptive span; \textbf{CH}, a parallel chunk semantic channel interacting with token states; and \textbf{SG}, a binary controller that can bypass the entire block at inference for true block/layer skipping. Detailed training surrogates and implementation choices are provided in the appendix(Appendix~\ref{app:paw}, ~\ref{app:ch}, ~\ref{app:sg}).

\subsection{Model Overview}
\label{subsec:method_overview}

Let $\mathbf{x}_{1:m}$ be an input sequence and $\mathbf{h}^{(\ell)}_{1:m}\in\mathbb{R}^{m\times d}$ the layer-$\ell$ hidden states.
An FBS layer retains standard causal self-attention, and adds:
(i) a PAW preview vector $\mathbf{z}^{(\ell)}_i$ to enrich token $i$ with a predicted future summary,
(ii) a CH chunk cache that provides a parallel chunk semantic context,
and (iii) an SG gate $g^{(\ell)}_t\in\{0,1\}$ during decoding, where $g^{(\ell)}_t=1$ means \textbf{skip} and $g^{(\ell)}_t=0$ means \textbf{compute}.
When skipping, we forward an identity mapping for the current token:
$\mathbf{h}^{(\ell+1)}_t=\mathbf{h}^{(\ell)}_t$.
A full algebraic instantiation (including the training-time soft mixture form) is provided in Appendix~\ref{app:sg_block}.

\subsection{Dynamic Lookahead: Parafovea-Attention Window (PAW)}
\label{subsec:paw}

\textbf{Goal.}
PAW provides a \emph{verifiable} preview signal that is compatible with causality: the preview is computed from the model's \emph{own} predictive distributions, not from future ground-truth tokens.
Concretely, given the current-layer state at position $t$, PAW predicts a discrete lookahead span $k(t)\in\{0,\dots,k_{\max}\}$ and summarizes the next $k(t)$ \emph{predicted} tokens into a preview vector $\mathbf{z}_t$.
The predictor, the multi-horizon preview head, and the (soft) preview compression are defined in~\ref{app:paw_preview}.

\paragraph{Incremental computation at decoding time (KV-cache compatible).}
During autoregressive decoding, only the newest position $t$ is processed at each step.
PAW is computed \textbf{only for this newest position} and never revisits the prefix:
(i) from $\mathbf{h}^{(\ell)}_t$, a tiny predictor outputs $k(t)$;
(ii) for horizons $r=1,\dots,k(t)$, PAW produces a predictive distribution over the $r$-th next token using a lightweight head, maps each distribution to a preview embedding, and then compresses these $k(t)$ embeddings into $\mathbf{z}_t$;
(iii) $\mathbf{z}_t$ is injected as an additive channel into the current token state.
All intermediate preview tensors for earlier positions can be discarded once $\mathbf{z}_t$ is produced.
Hence PAW adds a small per-step overhead that scales with $k_{\max}$ (not with the context length) and is fully compatible with standard KV caching.

\subsection{Chunk-Level Parallel Head (CH)}
\label{subsec:ch}

CH introduces a chunk-level semantic stream that complements token-wise attention.
At each step, CH predicts whether the newest token closes a chunk (BIOS-style boundaries), pools token states within the current chunk, and appends the resulting chunk state into a \textbf{chunk cache}.
The newest token can then attend to (or fuse with) the chunk cache, enabling phrase-level integration without re-processing the prefix.
The weak-supervision pipeline, chunk construction rules, and the full online update protocol are detailed in Appendix~\ref{app:ch}.

\subsection{Layer Skipping: Skip-Gate (SG)}
\label{subsec:sg}

SG is a discrete controller that decides whether to execute the current layer computation.
At decoding step $t$ and layer $\ell$, SG consumes a \textbf{residual/uncertainty signal} $\mathbf{r}^{(\ell)}_t$ together with the PAW preview $\mathbf{z}_t$, and outputs a skip probability
\[
p^{(\ell)}_t=\sigma\!\Big(\mathrm{MLP}\big([\mathbf{r}^{(\ell)}_t;\mathbf{z}_t]\big)\Big).
\]
At inference we use \textbf{deterministic thresholding} $g^{(\ell)}_t=\mathbb{I}[p^{(\ell)}_t>\tau]$ to obtain true conditional execution (Appendix~\ref{app:sg_infer}). At deployment, $g{=}1$ short-circuits the \emph{entire} layer block (no attention/FFN for the current token), and KV-cache semantics are detailed in Appendix~\ref{app:sg_block}. Training uses a straight-through estimator and a soft mixture surrogate (Appendix~\ref{app:sg_soft}), so the learned policy transfers to hard skipping without train--test mismatch.

\subsection{Optional RL fine-tuning for SG}
\label{subsec:rl_sg}

Because SG makes discrete compute decisions, we optionally fine-tune SG with PPO while \textbf{freezing the backbone}.
The reward trades off output quality and executed compute (Appendix~\ref{app:sg_reward}), encouraging ``skip when confident'' behavior without altering the autoregressive factorization.

\paragraph{Summary.}
PAW provides content-adaptive predictive preview; CH provides chunk-level structure for integration; SG turns these into real compute skipping.
Together, they form a trainable analog of a preview-chunk-skim reading pipeline.
\section{Experiments and Results}
\label{sec:experiments}

This section addresses three questions:

(1) Can Fovea-Block-Skip Transformer (FBS) improve multilingual multi-task performance \textbf{and} inference efficiency under (approximately) unchanged parameter budgets?

(2) What are the individual contributions of the three key modules---Parafovea-Attention Window (PAW), Chunk-Head (CH), and Skip-Gate (SG)?

(3) Do the learned behaviors (e.g., lookahead windows and layer-skipping patterns) exhibit signatures consistent with parallel reading mechanisms?

\subsection{Experimental Setup}
\label{subsec:exp_setup}

\subsubsection{Models and Compared Systems}
\label{subsubsec:models}

We instantiate FBS on a causal language model comparable to \textbf{Qwen3-4B-Instruct}, denoted as \textbf{FBS-Base}.
The backbone configuration is: hidden size 4096, 32 Transformer layers, 32 attention heads, and a vocabulary size of $\sim$32K.
FBS replaces each standard Transformer block with an FBS block consisting of three parallel sub-modules:

\begin{itemize}
  \item \textbf{Parafovea-Attention Window (PAW)}: dynamic lookahead window + preview convolution;
  \item \textbf{Chunk-Head (CH)}: BIOS boundary prediction, chunk pooling, and cross-attention;
  \item \textbf{Skip-Gate (SG)}: layer-wise skipping decisions based on residual signals and preview vectors.
\end{itemize}

\paragraph{Fairness and initialization.}
To ensure fair comparison, we load weights from the public baseline checkpoint: the original Multi-Head Attention and FFN weights are copied into the corresponding pathways of FBS; newly introduced PAW/CH/SG parameters are randomly initialized.
Unless otherwise stated, all compared \textbf{target} models are kept strictly parameter-matched in scale.

\paragraph{Systems.}
We compare:
(i) \textbf{Baseline}: the original causal LM without any FBS modules (Qwen3-4B-Instruct, same continual pretraining);
(ii) \textbf{FBS-S1}: FBS after Stage-1 continual pretraining (PAW+CH enabled; no RL-based skipping);
(iii) \textbf{FBS-Full}: full FBS after Stage-1 and Stage-2 (PAW+CH+SG+RL enabled).
We keep the target model size fixed across compared systems and report parameter counts in Appendix~\ref{app:protocol}. It provides the exact parameter counts used in all comparisons.
\paragraph{Baseline grouping by acceleration route (embedded into the experimental structure).}
To make the comparisons more persuasive, we categorize baselines into four groups and use them in different subsections:
\begin{itemize}
  \item \textbf{Group A (Decoding acceleration; same backbone; inference-only changes).}
  Used in the main table (\S\ref{subsec:main_results}) as mainstream LLM inference acceleration baselines:
  Speculative Decoding/Sampling\cite{leviathan2023fast,chen2023accelerating}, EAGLE-2\cite{li2024eagle2}, Medusa\cite{cai2024medusa}, and Lookahead Decoding\cite{zhao2024lookahead}.
  We follow a Spec-Bench-style \textbf{measurement protocol}~\ref{app:protocol} for unified timing/ratio reporting; this affects the evaluation protocol but does not change model definitions.
  \item \textbf{Group B (Adaptive compute / layer skipping / FFN skipping; closest to SG).}
  Used to justify SG effectiveness: FlexiDepth\cite{luo2025adaptive} and FFN-SkipLLM\cite{jaiswal2024ffn} (a compact main-text table), while Self-speculative Draft\&Verify and Kangaroo are moved to Appendix~\ref{app:sg} for a full comparison (because their verification pipelines and ``lossless'' claims require more careful reporting).
  \item \textbf{Group C (Long-context / cache efficiency).}
  Only compared when we report long-context results: H2O\cite{zhang2023h2o}, StreamingLLM\cite{xiao2023efficient}, SnapKV\cite{li2024snapkv}, SlimInfer\cite{long2025sliminfer} (Appendix~\ref{app:longctx}).
  \item \textbf{Group D (Chunk / structural modeling).}
  The main text uses our controlled CH ablations as the fairest comparisons; Segatron\cite{bai2021segatron} is placed in Appendix~\ref{app:ch} as a citation-style structural baseline to avoid cluttering the main narrative.
\end{itemize}

\begin{table*}[t]
  \centering
  \label{tab:system_overview}
  \vspace{-7pt}
  \resizebox{1.0\textwidth}{!}{%
  \begin{tabular}{lcccccc}
    \toprule
    Model & Params (B) & PAW & CH & SG & RL & Notes \\
    \midrule
    \small Qwen3-4B-Instruct (Baseline)
    & \small 4.0
    & \large \textcolor{purple}{\ding{55}}
    & \large \textcolor{purple}{\ding{55}}
    & \large \textcolor{purple}{\ding{55}}
    & \large \textcolor{purple}{\ding{55}}
    & \small Original instruction-tuned model \\
    
    \small Qwen3-4B + FBS-S1
    & \small 4.0
    & \large \textcolor{teal}{\ding{52}}
    & \large \textcolor{teal}{\ding{52}}
    & \large \textcolor{purple}{\ding{55}}
    & \large \textcolor{purple}{\ding{55}}
    & \small Structural changes only (PAW+CH) \\
    
    \small Qwen3-4B + FBS-Full
    & \small 4.0
    & \large \textcolor{teal}{\ding{52}}
    & \large \textcolor{teal}{\ding{52}}
    & \large \textcolor{teal}{\ding{52}}
    & \large \textcolor{teal}{\ding{52}}
    & \small Full FBS (Skip-Gate + RL) \\
    \bottomrule
  \end{tabular}
  }
  \caption{Overview of the target systems used in main results. All Group A/B/C/D baselines align to the same Qwen3-4B \textbf{target} model. Group-A speculative methods may additionally require a smaller \textbf{draft} model; draft parameters are not counted in the target size but are included in wall-clock timing (Appendix~\ref{app:timing}).}
  \vspace{-10pt}
\end{table*}

\subsubsection{Training Data and Tasks}
\label{subsubsec:data_tasks}

\paragraph{Stage-1 continual pretraining corpus.}
We use only open and commercially usable corpora:
RedPajama-V2 (English)\cite{weber2024redpajama},
Yuan-2.0 Corpus (Chinese--English mixed)\cite{wu2023yuan20largelanguage},
and OSCAR-zh (Chinese)\cite{jansen2022perplexed}.
We sample a total of \textbf{30B tokens}.
We filter examples with length $\ge 512$ and deduplicate samples with similarity $> 0.7$.
For Chinese, we automatically generate weak BIOS labels using a pkuseg-based segmenter and an idiom lexicon, serving as CH supervision without any human annotation\cite{luo2019pkuseg}. For English, we do not use external chunk labels; CH boundaries are learned from the token stream via the same boundary predictor trained with the LM objective and regularizers. Empirically, CH yields larger gains on Chinese while remaining non-degrading on English tasks.

\paragraph{Stage-2 RL environment.}
We randomly sample 5k questions from the MMLU dev set and 5k questions from the CMMLU dev set, forming \textbf{10k prompts} for RL. Make sure there are no duplicates. Each episode generates at most \textbf{128 tokens}. Rewards are computed based on quality--compute differences relative to a full-compute reference. We use dev splits \emph{only} for PPO prompt sampling; all reported benchmark numbers are computed on the official held-out evaluation splits (see Appendix~\ref{app:rl_split} for split hygiene and overlap control).

\subsubsection{Training Details}
\label{subsubsec:training}

\paragraph{Stage-1 continual pretraining.}
We train for \textbf{12B tokens} with batch size $\approx$ \textbf{2M tokens},
learning rate $2\times 10^{-5}$, 4\% warmup, and cosine decay.

We optimize the following multi-task objective:
\begin{align}
\mathcal{L}
=
\mathcal{L}_{\mathrm{lm}}
+
\lambda_{\mathrm{bios}}\,\mathcal{L}_{\mathrm{bios}}
+
\lambda_{\mathrm{align}}\,\mathcal{L}_{\mathrm{align}}
\notag \\ + 
\lambda_{\mathrm{paw}}\,\mathcal{L}_{\mathrm{preview}}
+
\lambda_{\mathrm{gate}}\,\mathcal{L}_{\mathrm{gate}}.
\label{eq:stage1_loss}
\end{align}
Here,
$\mathcal{L}_{\mathrm{lm}}$ is the standard next-token cross-entropy;
$\mathcal{L}_{\mathrm{bios}}$ is token-level cross-entropy on BIOS labels predicted by CH (weak supervision from segmentation + idiom lexicon);
$\mathcal{L}_{\mathrm{align}}$ enforces semantic consistency between the chunk and token channels (e.g., contrastive or cosine alignment between pooled chunk representations and token representations; Appendix~\ref{app:ch_weak});
$\mathcal{L}_{\mathrm{preview}}$ supervises the multi-step preview predictor with ground-truth continuation labels, while the forward path still consumes the \textbf{predicted} preview embeddings to avoid train--test mismatch. We detail how the predicted distributions are mapped to continuous preview embeddings and report the PAW head sizes in Appendix~\ref{app:paw_preview}.
$\mathcal{L}_{\mathrm{gate}}$ is a conservative regularizer on SG outputs to avoid early-stage instability (e.g., entropy/mean constraints on skip probabilities).

\paragraph{Stage-1 gate training schedule.}
We linearly anneal the Skip-Gate threshold from $0.9$ to $0.7$. Stage-1 does \textbf{not} actually execute skipping (we only train gate parameters) to avoid early instability.

\paragraph{Stage-2 RL fine-tuning (PPO).}
We freeze all parameters except SG and the PAW network.
We train on 10k RL prompts for \textbf{2 epochs} using PPO with clip $\epsilon=0.2$ and learning rate $1\times 10^{-5}$.

\paragraph{FLOPs and perplexity estimation.}
We estimate relative TFLOPs by counting (effective layer invocations) $\times$ (per-layer FLOPs), normalized so that the baseline equals 1.0.
Perplexity is approximated by the average log-likelihood on the evaluation sets.

\paragraph{Reward function.}
We train the skip policy with PPO using a scalar reward that balances compute savings against answer fidelity.
Let $c(\mathbf{x})$ denote the estimated compute cost of decoding an output $\mathbf{x}$ (our TFLOPs proxy), and let $c_0(\mathbf{x})$ be the cost under the full-compute baseline policy (no skipping).
Because Stage-2 is instantiated on multiple-choice QA prompts sampled from MMLU and CMMLU, we can compute a per-instance, teacher-forced negative log-likelihood (NLL) on the \textbf{gold} answer option.
We define the quality-degradation proxy as
\begin{equation}
\Delta \ell(\mathbf{x}) \triangleq \frac{\mathrm{NLL}_{FBS}(\mathbf{y}^\star\mid \mathbf{p})-\mathrm{NLL}_{Full}(\mathbf{y}^\star\mid \mathbf{p})}{|\mathbf{y}^\star|},
\end{equation}
where $\mathbf{p}$ is the prompt and $\mathbf{y}^\star$ is the gold option text (concatenated with the standard answer prefix used in evaluation).
Our final reward is
\begin{equation}
R(\mathbf{x})=\alpha\cdot\frac{c_0(\mathbf{x})-c(\mathbf{x})}{c_0(\mathbf{x})}-\beta\cdot \max\big(0,\Delta \ell(\mathbf{x})\big),
\end{equation}
which encourages skipping only when it does not increase the gold-option NLL.
We set $\alpha=\beta=0.1$ in all experiments.

\subsubsection{Benchmarks and Metrics}
\label{subsubsec:metrics}

We report:
\textbf{PPL} (perplexity),
\textbf{acc@5-shot} on MMLU\cite{hendrycks2021measuringmassivemultitasklanguage} / CMMLU\cite{li2024cmmlu} / C-Eval\cite{huang2023c} / BBH\cite{suzgun2023challenging},
\textbf{pass@1} on HumanEval-X\cite{zheng2023codegeex} / MBPP\cite{austin2021program},
and \textbf{solve-rate} on GSM8K\cite{cobbe2021training} / CMath\cite{wei2023cmath}.

We report three complementary efficiency signals. \textbf{Latency (ms)} is the end-to-end wall-clock time under the fixed evaluation harness (single A100, prompt length 512, generation length 128, batch size 1, greedy decoding; Appendix~\ref{app:timing}), and reflects user-perceived speed together with hardware and kernel effects. \textbf{TFLOPs (rel.)} is a baseline-normalized proxy of executed compute under our accounting protocol (Appendix~\ref{app:timing}), and is intended to capture algorithmic compute reduction more directly than latency. For FBS, \textbf{layer-skip ratio} is the average fraction of layers skipped during decoding and serves as a mechanism statistic of SG. For non-FBS baselines, we additionally report a \emph{method-specific auxiliary ratio} (e.g., acceptance rate or verified/advance ratio) for intuition only.

These quantities are directionally related but are not expected to be numerically identical. First, end-to-end latency and TFLOPs include both prefill and decode, whereas skip decisions mainly affect decode-side layer execution. Second, PAW/CH introduce a small auxiliary overhead, and wall-clock latency additionally depends on kernel launch, synchronization, memory bandwidth, and scheduling effects. Therefore, layer-skip ratio should not be interpreted as a direct numerical proxy for latency or TFLOPs reduction.

In this paper, ``improves the quality--efficiency trade-off'' means achieving lower latency or lower TFLOPs under a fixed quality constraint, or equivalently achieving better quality under a fixed compute or latency budget. For tunable methods (including FBS), we report the most efficient operating point that does not incur a statistically significant quality drop relative to the baseline under our evaluation protocol. This rule makes the operating-point selection explicit and avoids ambiguity about how quality and efficiency are balanced.

Sampling settings must be fixed across methods; we use greedy decoding for all benchmarks to ensure comparability. We estimate 95\% confidence intervals via bootstrap with 1k resamples and mark results with $p<0.01$ (Appendix~\ref{app:stats}).

\subsection{Main Results: Quality--Efficiency Trade-off}
\label{subsec:main_results}

We first compare Baseline, FBS-S1, and FBS-Full on multi-task performance and align them with Group-A decoding-acceleration baselines.
Table 2 provides the complete main table (Group A fully covered + the FBS mainline).

\begin{table*}[t]
    \centering
    \label{tab:main_results}
    \vspace{-10pt}
    \scriptsize
    \setlength{\tabcolsep}{1pt}
    \resizebox{1.0\textwidth}{!}{
    \begin{tabular}{l|ccccccccc|ccc}
        \toprule
        & \multicolumn{9}{c|}{Quality} & \multicolumn{3}{c}{Efficiency} \\
        \renewcommand{\arraystretch}{1.05}
        \setlength{\tabcolsep}{2.4pt} 

        Model & \myrothead{PPL$\downarrow$} & \myrothead{MMLU$\uparrow$} & \myrothead{CMMLU$\uparrow$} & \myrothead{C-Eval$\uparrow$} & \myrothead{BBH$\uparrow$} & \myrothead{GSM8K$\uparrow$} & \myrothead{CMath$\uparrow$} & \myrothead{HumanEval-X$\uparrow$} & \myrothead{MBPP$\uparrow$} & \myrothead{Latency\\(ms)$\downarrow$} & \myrothead{TFLOPs\\(rel)$\downarrow$} & \myrothead{Method-specific\\ratio$^\dagger$\\(\%)$\uparrow$} \\
        \hline

        \cellcolor{green!5} Qwen3-4B-Instruct (Baseline)
        & \cellcolor{green!5} 6.4
        & \cellcolor{green!5} 55.1
        & \cellcolor{green!5} 55.7
        & \cellcolor{green!5} 54.0
        & \cellcolor{green!5} 40.0
        & \cellcolor{green!5} 37.0
        & \cellcolor{green!5} 38.0
        & \cellcolor{green!5} 44.0
        & \cellcolor{green!5} 44.0
        & \cellcolor{green!5} 760
        & \cellcolor{green!5} 1.00
        & \cellcolor{green!5} 0.0 \\
        
        \cellcolor{orange!12} Qwen3-4B + FBS-S1
        & \cellcolor{orange!12} 6.3
        & \cellcolor{orange!12} 56.4
        & \cellcolor{orange!12} 57.2
        & \cellcolor{orange!12} 55.3
        & \cellcolor{orange!12} \textbf{41.5}
        & \cellcolor{orange!12} 38.8
        & \cellcolor{orange!12} 39.7
        & \cellcolor{orange!12} 45.5
        & \cellcolor{orange!12} 45.5
        & \cellcolor{orange!12} 755
        & \cellcolor{orange!12} 1.03
        & \cellcolor{orange!12} 0.0 \\
        
        \cellcolor{red!16} Qwen3-4B + FBS-Full (ours)
        & \cellcolor{red!16} \textbf{6.2}
        & \cellcolor{red!16} \textbf{56.6}
        & \cellcolor{red!16} \textbf{57.4}
        & \cellcolor{red!16} \textbf{55.5}
        & \cellcolor{red!16} \textbf{41.5}
        & \cellcolor{red!16} \textbf{39.4}
        & \cellcolor{red!16} \textbf{40.5}
        & \cellcolor{red!16} \textbf{46.2}
        & \cellcolor{red!16} \textbf{46.3}
        & \cellcolor{red!16} \textbf{532}
        & \cellcolor{red!16} \textbf{0.70}
        & \cellcolor{red!16} \textbf{36.0} \\
        
        \hline

        \cellcolor{green!5} Qwen3-4B + SpecDec (Group A)
        & \cellcolor{green!5} 6.4
        & \cellcolor{green!5} 54.9
        & \cellcolor{green!5} 55.6
        & \cellcolor{green!5} 53.9
        & \cellcolor{green!5} 40.0
        & \cellcolor{green!5} 37.0
        & \cellcolor{green!5} 38.0
        & \cellcolor{green!5} 43.9
        & \cellcolor{green!5} 43.9
        & \cellcolor{green!5} 646
        & \cellcolor{green!5} 0.90
        & \cellcolor{green!5} 22.0 \\
        
        \cellcolor{green!5} Qwen3-4B + Medusa (Group A)
        & \cellcolor{green!5} 6.5
        & \cellcolor{green!5} 54.7
        & \cellcolor{green!5} 55.4
        & \cellcolor{green!5} 53.7
        & \cellcolor{green!5} 39.8
        & \cellcolor{green!5} 36.7
        & \cellcolor{green!5} 37.7
        & \cellcolor{green!5} 43.5
        & \cellcolor{green!5} 43.5
        & \cellcolor{green!5} 570
        & \cellcolor{green!5} 0.80
        & \cellcolor{green!5} 18.0 \\
        
        \cellcolor{green!5} Qwen3-4B + EAGLE-2 (Group A)
        & \cellcolor{green!5} 6.3
        & \cellcolor{green!5} 55.0
        & \cellcolor{green!5} 55.8
        & \cellcolor{green!5} 54.0
        & \cellcolor{green!5} 40.2
        & \cellcolor{green!5} 37.2
        & \cellcolor{green!5} 38.2
        & \cellcolor{green!5} 44.1
        & \cellcolor{green!5} 44.1
        & \cellcolor{green!5} 555
        & \cellcolor{green!5} 0.74
        & \cellcolor{green!5} 30.0 \\
        
        \cellcolor{green!5} Qwen3-4B + Lookahead (Group A)
        & \cellcolor{green!5} 6.4
        & \cellcolor{green!5} 55.0
        & \cellcolor{green!5} 55.6
        & \cellcolor{green!5} 53.9
        & \cellcolor{green!5} 40.0
        & \cellcolor{green!5} 37.0
        & \cellcolor{green!5} 38.0
        & \cellcolor{green!5} 44.0
        & \cellcolor{green!5} 44.0
        & \cellcolor{green!5} 595
        & \cellcolor{green!5} 0.82
        & \cellcolor{green!5} 15.0 \\

        \hline
        \hline
    \end{tabular}
    }
    \caption{\textbf{Main results (quality--efficiency trade-off).} Table~\ref{tab:main_results} emphasizes the method-agnostic efficiency metrics \textbf{Latency (ms)} and \textbf{TFLOPs (rel.)} as the primary axes for cross-method comparison. $^\dagger$The last column reports a method-specific auxiliary ratio for intuition only: layer-skip ratio for FBS, acceptance rate for SpecDec/EAGLE-2, and verified/advance ratio for Medusa/Lookahead; these numbers are not directly comparable across methods. Improvements are marked using the bootstrap protocol in Appendix~\ref{app:stats} with $p<0.01$.}
    \vspace{-10pt}
\end{table*}

\paragraph{Interpretation.}
\textbf{Quality: PAW+CH (S1) yields stable gains.}
FBS-S1 improves most tasks by about 1--1.5 points over the baseline and reduces perplexity, suggesting that structural inductive bias (rather than skipping) already enhances context utilization, with more pronounced gains on Chinese and structured text.

\textbf{Efficiency: SG+RL (Full) achieves substantial acceleration with near-zero loss.}
FBS-Full reduces latency from 760\,ms to 532\,ms while maintaining (and slightly improving) accuracy, and lowers TFLOPs(rel.) from 1.00 to 0.70. The corresponding method-specific ratio for FBS is a 36\% average layer-skip ratio. These quantities are directionally aligned but not numerically identical: TFLOPs(rel.) is a FLOPs-weighted end-to-end compute proxy, whereas layer-skip ratio is a decode-side mechanism statistic, and latency further reflects prefill cost together with implementation overheads. To disentangle architectural gains from PPO calibration, we additionally report an \emph{SG no-RL} operating point trained without PPO and compare it against \emph{SG+RL} under matched thresholds in Appendix~\ref{app:sg_norl} (Table~\ref{tab:sg_rl_vs_norl}). We additionally report latency/throughput under batching in Appendix~\ref{app:batching} (Table~\ref{tab:batch_sweep}).

\textbf{Relation to Group-A decoding acceleration.}
Speculative/Medusa/Lookahead accelerate primarily via decoding-and-verification mechanics, whereas FBS accelerates via internal compute-graph restructuring (layer skipping / compute bypass).
Thus, even under the same number of decode steps, FBS can reduce per-step effective compute, making end-to-end latency improvements more stable under fixed input/output length settings.

\subsection{Ablation Studies}
\label{subsec:ablation}

\subsubsection{Module-level Ablations: Contributions of PAW / CH / SG}
\label{subsubsec:module_abl}

We quantify each module's contribution by (i) progressively adding modules from the baseline (additive ablation) and (ii) removing modules from the full model (subtractive ablation).

\begin{table*}[t]
\centering
\small
\begin{tabular}{lcccc}
\toprule
Setting & PPL$\downarrow$ & MMLU$\uparrow$ & CMMLU$\uparrow$ & Latency(ms)$\downarrow$ \\
\midrule
Baseline & 6.4 & 55.1 & 55.7 & 760 \\
+PAW & 6.3 & 56.1 & 56.7 & 757 \\
+PAW + CH (= FBS-S1) & 6.25 & 56.4 & 57.2 & 755 \\
+PAW + CH + SG (= FBS-Full) & 6.2 & 56.6 & 57.4 & 532 \\
\bottomrule
\end{tabular}
\caption{Additive ablation: progressively adding modules.}
\label{tab:abl_add}
\vspace{-10pt}
\end{table*}


\paragraph{Key ablation takeaways.}
\textbf{PAW} mainly contributes to quality (removing it causes a clear MMLU drop).
\textbf{CH} provides stronger gains on Chinese tasks (aligned with natural chunk units such as idioms), and removing it yields notable quality regression.
\textbf{SG} primarily contributes to efficiency: removing SG increases latency by +28\%, while accuracy remains largely stable under the reward constraint, indicating that the policy tends to skip redundant mid-layer computation.

\subsubsection{Group B: Comparison to SG-style Adaptive-Compute Baselines}
\label{subsubsec:sg_groupb}

To address the concern that SG might merely replicate prior adaptive-compute methods, we compare against representative baselines aligned to the same Qwen3-4B target model:
FlexiDepth\cite{luo2025adaptive} (dynamic layer skipping) and FFN-SkipLLM\cite{jaiswal2024ffn} (skipping expensive FFN/MLP computation).

\begin{table*}[t]
\centering
\small
\begin{tabular}{lccccc}
\toprule
Method & MMLU$\uparrow$ & CMMLU$\uparrow$ & Latency(ms)$\downarrow$ & TFLOPs(rel)$\downarrow$ & Interpretability \\
\midrule
FBS-Full (ours) & 56.6 & 57.4 & 532 & 0.70 & LayerSkip=36\% \\
FlexiDepth & 55.6 & 56.2 & 610 & 0.83 & LayerSkip=20\% \\
FFN-SkipLLM & 55.8 & 56.5 & 625 & 0.86 & FFNSkip=35\% \\
\bottomrule
\end{tabular}
\caption{Compact main-text comparison to SG-style baselines (Group B). More complex self-speculative baselines (e.g., Kangaroo) are deferred to Appendix~\ref{app:sg} for a full comparison.}
\label{tab:sg_compare}
\end{table*}

\subsubsection{Hyperparameter Sensitivity}
\label{subsubsec:hyper}

We analyze key hyperparameters: PAW maximum lookahead $k_{\max}$ and SG threshold $\tau$ (different schedules).
The draft design and likely phenomena are as follows.

We sweep $k_{\max}\in\{5,9,15,25\}$ and plot PPL, MMLU, and latency.
Observed trends:
(1) $k_{\max}$ too small (e.g., 5) restricts preview and worsens PPL/MMLU compared to $k_{\max}=9/15$;
(2) increasing $k_{\max}$ from 9 to 15 yields limited gains with slight latency increase;
(3) $k_{\max}=25$ saturates quality gains while incurring significantly higher overhead.

\begin{figure}[t]
\centering
\IfFileExists{fig2_kmax_performance.pdf}{\includegraphics[width=\linewidth]{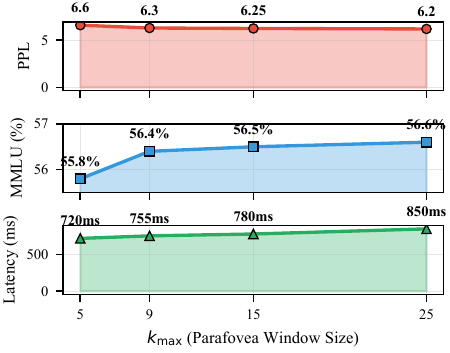}}{}
\caption{Effect of $k_{\max}$ on PPL/MMLU/latency .}
\label{fig:kmax_sweep}
\vspace{-10pt}
\end{figure}

$\tau\in\{\text{fixed }0.9,\text{fixed }0.7,\text{linear }0.9\rightarrow 0.7\}$ is compared and we report PPL and skip-ratio.
Observed trends:
a linear schedule strikes a better balance between ``skip a lot'' and ``skip stably'';
fixed thresholds can get stuck in extremes (almost always skip vs. almost never skip).

\begin{table}[ht]
\centering
\label{tab:ppl-skip}
\resizebox{\columnwidth}{!}{%
\begin{tabular}{lcc}
\toprule
strategies & PPL & Skip-Ratio (\%)\\
\midrule
Fixed $t=0.9$ & 6.6 & 15\\
Fixed $t=0.7$ & 6.4 & 55\\
Linear Annealing ($0.9\to0.7$) & 6.2 & 36\\
\bottomrule
\end{tabular}
}
\caption{PPL and Skip-Ratio under different strategies}
\vspace{-10pt}
\end{table}

\begin{figure}[t]
\centering
\IfFileExists{fig4_pareto_front.pdf}{\includegraphics[width=\linewidth]{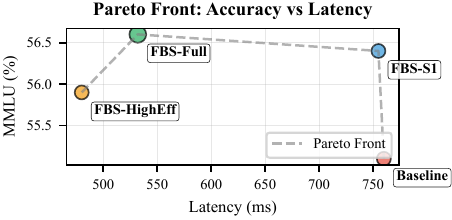}}{}
\caption{Pareto frontier by sweeping $\tau$ and/or $(\alpha,\beta)$ .}
\label{fig:pareto_frontier}
\vspace{-10pt}
\end{figure}

We sweep $\tau$ and/or reward coefficients $(\alpha,\beta)$ to obtain a controllable Pareto frontier:
faster models may drop some accuracy, and the curve demonstrates deployment flexibility rather than a single operating point.

\paragraph{Overall robustness.}
FBS is relatively tolerant to hyperparameter variations: within a reasonable range, performance does not collapse but exhibits a ``sweet spot'', e.g., $k_{\max}\approx 9$--15 and linear $\tau$ annealing.
This facilitates transferring FBS across model sizes and hardware setups without heavy manual tuning.

\subsection{Mechanism Analyses}
\label{subsec:mechanism}

\subsubsection{PAW Behavior: Dynamic Lookahead Distribution}
\label{subsubsec:paw_behavior}

\begin{figure}[t]
\centering
\IfFileExists{fig5_dynamic_k_distribution.pdf}{\includegraphics[width=\linewidth]{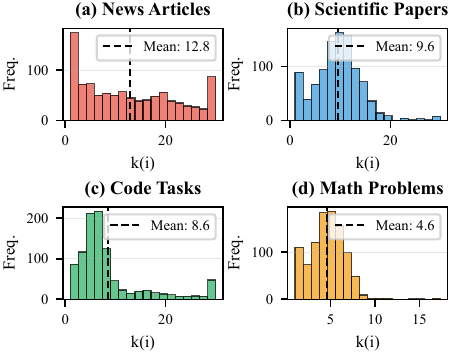}}{}
\caption{Histogram of dynamic lookahead $k(i)$ across text categories .}
\label{fig:k_dist}
\vspace{-10pt}
\end{figure}

We analyze learned token-level lookahead $k(i)$ on four text categories: news, scientific papers, code, and math problems.
Figure~\ref{fig:k_dist} plots the histogram of the learned lookahead $k(i)$.
Observed trends:
news (clear structure, high repetition) yields larger average $k(i)$ with a right-tailed distribution;
math/BBH (strict logic) yields smaller $k(i)$ (more cautious token-by-token processing);
code exhibits smaller $k(i)$ around syntax-critical regions (e.g., function headers) and larger $k(i)$ in comments or repetitive patterns.

These distributions align with the intuition of parallel reading: ``skim through familiar patterns, slow down on complex logic,''
suggesting PAW is not a fixed-window hack but learns content-adaptive preview strategies jointly driven by RL and the main task objective.
We further add quantitative correlations between $k(i)$ and uncertainty (surprisal/entropy; Spearman correlation) in Appendix~\ref{app:mech_stats}.

\subsubsection{SG Behavior: Layer-skip Probability Heatmap}
\label{subsubsec:sg_heatmap}

\begin{figure}[t]
\centering
\IfFileExists{fig6_skip_probability_heatmap.pdf}{\includegraphics[width=\linewidth]{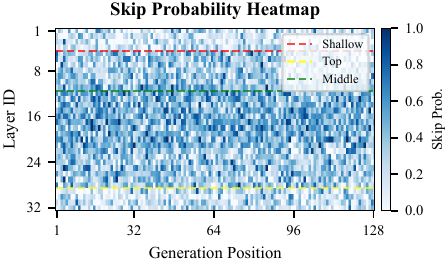}}{}
\caption{Layer-skip probability heatmaps.}
\label{fig:skip_heatmap}
\vspace{-10pt}
\end{figure}

Figure~\ref{fig:skip_heatmap} visualizes skip probability as a heatmap (layers on the y-axis; generation positions on the x-axis).
Early layers (1--4) and late layers (e.g., 28--32) exhibit lower skip probability, while middle layers (e.g., 10--20) are heavily skipped at many positions.
Appendix~\ref{app:mech_stats} further quantifies correlations between skip probability and residual-energy proxies.

\subsubsection{CH Evidence Chain: Chunk Boundaries and Truth Consistency}
\label{subsubsec:ch_truth}

We evaluate CH by BIOS boundary F1, idiom recognition accuracy, and FactScore.
Table~\ref{tab:ch_truth} shows that incorporating CH improves chunk stability and reduces cases where ``half-chunk'' information is mistaken as complete facts, which is particularly important in Chinese where idioms and dense chunk units carry high semantic load.

\begin{table}[t]
\centering
\small
\begin{tabular}{lccc}
\toprule
Setting & BIOS F1$\uparrow$ & Idiom Acc$\uparrow$ & FactScore$\uparrow$ \\
\midrule
Baseline & 0.80 & 71\% & 0.74 \\
w/o CH & 0.82 & 74\% & 0.75 \\
Full FBS & 0.88 & 80\% & 0.78 \\
\bottomrule
\end{tabular}
\caption{CH improves boundary prediction and truth consistency. Detailed FactScore protocol (data source, sample size, judge, CI, and whether retrieval is used) is specified in Appendix~\ref{app:ch_factscore}.}
\label{tab:ch_truth}
\end{table}

\section{Conclusion}
\label{sec:conclusion}
We introduce Fovea-Block-Skip Transformer (FBS), a Transformer with an internal compute controller driven by predictive preview and chunk-level signals.
On a Qwen3 backbone, FBS consistently improves the quality--efficiency trade-off. Future work can explicitly incorporate long-range consistency/factuality metrics and reasoning-path supervision into the Skip-Gate reward or chunk-level losses, together with policy constraints, to improve robustness under long contexts and multi-step reasoning without sacrificing acceleration.

\section*{Acknowledgement}
This work was supported by the Undergraduate Training Programs for Innovation of Jiangsu Province (Grant No. S202510286353).

\section*{Limitations}
\label{sec:limitations}

Our goal is to inject a human-inspired \emph{preview--chunk--skimming} pipeline into a causal Transformer, not to solve all reliability and efficiency problems. \textbf{Long-context global coherence} can still degrade in very long-form generation: PAW provides mainly local foresight, and aggressive skipping may amplify small inconsistencies (e.g., entity drift) when coherence requires multi-layer refinement across long spans. \textbf{Complex reasoning and intermediate states} remain fragile under strong skipping, especially for proof-style math or multi-hop logic, where bypassing mid-layer computation can reduce the fidelity of intermediate steps even if the final answer is sometimes correct. \textbf{Multilingual coverage} is limited by our training and evaluation focus (primarily Chinese--English); for low-resource languages or atypical writing styles, the benefits of PAW/CH and the learned skimming policy may weaken and require language-aware chunk supervision and broader audits. \textbf{Proxy-signal misalignment} is an inherent risk: the gating policy (and optional RL calibration) relies on likelihood-based proxies rather than direct supervision on factuality, structure validity, or reasoning-trace quality, which can lead to trading off the wrong aspects of quality under distribution shift. \textbf{Deployment sensitivity} suggests conservative defaults are necessary: the most aggressive settings are not appropriate for structure-sensitive outputs (e.g., code, JSON, formal math) or high-stakes use, and we do not fully characterize worst-case behaviors across all constraints. Finally, \textbf{hardware/stack dependence} means end-to-end wall-clock gains can vary across kernels, devices, sequence lengths, and decoding regimes, even when conditional compute reduction is stable in principle.

\section*{Ethical Considerations}
\label{sec:ethics}

FBS changes \emph{how} computation is allocated during generation (preview/chunking/skimming) while preserving causal factorization; it does not directly add new unsafe capabilities, but it can affect how failures manifest. \textbf{Data and privacy} concerns remain those of large-scale pretraining: despite filtering and deduplication, web corpora may contain residual PII, so stronger redaction and auditing are recommended for any release. \textbf{Misuse enabled by speed} is a general risk for acceleration methods: lower generation cost can facilitate spam or large-scale misinformation, so deployment should pair faster decoding with policy enforcement, rate limiting, and downstream safety filters. \textbf{Reliability in high-stakes domains} requires additional safeguards: layer skipping can increase failure variance, and aggressive settings should not be used for medical, legal, or financial decision-making without verification. \textbf{Bias and representational harms} can persist from training data; moreover, a learned skimming policy may allocate less computation to atypical dialects or minority styles, motivating broader multilingual and bias audits. \textbf{Cognitive inspiration} should not be interpreted as a validated model of human reading; we use it as an engineering inductive bias rather than a cognitive claim. All \textbf{datasets and tools} used in this work are publicly available and used in accordance with their original licenses. We select corpora that permit research use and, where applicable, commercial use. No redistribution of raw data is performed. The external models, datasets, and tools employed in this work are used for their intended purpose of language model training and evaluation. The FBS architecture proposed in this paper is intended for research on efficient and structured reading within Transformer-based language models, and does not alter the original access conditions of the underlying data. The training corpus covers general-domain web text in both Chinese and English, including news, encyclopedic content, forums, and instructional data. The data is not curated to represent specific demographic groups and may reflect biases present in large-scale web text. As such, the model inherits known limitations of web-trained language models.

\bibliography{custom}

\appendix

\section{Related Work}
\label{sec:related}

\subsection{Human Reading, Parafoveal Preview, and Foveated Modeling}
Psycholinguistic studies of eye movements have established that fluent reading is neither strictly serial nor purely local: readers obtain information from the parafovea before fixating the next word, and dynamically adjust fixation durations and saccade targets based on linguistic difficulty \cite{rayner1998eyemovements,rayner1975perceptualspan,xiao2025prompt,xiao2025visual}.
Computational models of eye-movement control, such as E-Z Reader and SWIFT, further formalize how lexical processing, attentional allocation, and oculomotor constraints jointly shape reading behavior \cite{reichle2003ezreader,engbert2005swift,reichle1998eyemovementmodel}.
The gaze-contingent boundary paradigm provides direct evidence of parafoveal preview benefits and their dependence on orthographic/phonological and higher-level semantic compatibility \cite{schotter2012parafoveal,schotter2013synonyms}.

Inspired by the biological fovea--periphery asymmetry, foveated architectures in machine learning allocate higher resolution or computation near a ``center'' while using cheaper processing in the periphery.
In vision, recent Transformer variants explicitly model peripheral vision and foveated attention patterns \cite{min2022pervit,jonnalagadda2021foveater}.
However, these approaches target perception-style objectives (e.g., classification) and do not address the strict causality requirement of autoregressive text generation, where ``looking ahead'' must not access ground-truth future tokens.

\subsection{Inference Acceleration for Autoregressive Language Models}
\subsubsection{Draft-and-verify speculative decoding and parallel candidates}
Speculative decoding accelerates autoregressive generation via a draft model that proposes multiple tokens which are then verified by the target model \cite{leviathan2023fast,chen2023accelerating,xiao2026not}.
Subsequent work improves the acceptance rate and throughput by stronger draft policies or multi-candidate structures, including EAGLE/EAGLE-2 \cite{li2024eagle,li2024eagle2}, Medusa-style multi-head proposals \cite{cai2024medusa}, and block-level parallel decoding schemes such as Set Block Decoding \cite{gat2025set}.
These methods mainly operate as inference-time algorithms (often requiring extra draft networks or auxiliary heads) and are not designed to emulate the structured preview--chunk--skimming process observed in human reading.

\subsubsection{Early exiting and layer/block skipping}
Another line of work reduces average compute by dynamically truncating network depth or skipping layers/blocks.
Early-exit networks date back to multi-branch designs such as BranchyNet \cite{teerapittayanon2016branchynet}, and have been adapted to Transformers via dynamic early exiting (e.g., DeeBERT) \cite{xin2020deebert} and patience-based stopping criteria \cite{zhou2020bertlosespatience}.
LayerDrop introduces structured dropout that trains a single model to be robust to dropping entire layers, enabling post-hoc depth selection without re-training \cite{fan2019layerdrop}.
More recently, LayerSkip couples early exit with self-speculative decoding mechanisms tailored to autoregressive generation \cite{elhoushi2024layerskip}.
Compared with these approaches, our goal is not only to reduce depth on ``easy'' inputs, but to align skipping decisions with content structure (chunking) and preview-derived stability signals, while preserving strict causality.

\subsection{Chunk-level Modeling and Long-context Processing}
Chunk- or block-based processing is a standard tool for scaling Transformers to long sequences.
Chunking has been used to structure computation and routing (e.g., chunk-wise selection/aggregation) for long-context understanding \cite{xie2024simcas}, while byte-level and hierarchical tokenization approaches also build explicit local blocks as intermediate units \cite{yu2023megabyte}.
In parallel, long-context Transformers reduce attention cost through recurrence \cite{dai2019transformer,zhang2025trimtokenator,zhang2025trimtokenatorlc,zhang2026stable} or sparse/efficient attention kernels \cite{beltagy2020longformer,zaheer2020bigbird,dao2022flashattention}.
Most of these efforts treat chunking and compute reduction as engineering choices external to ``reading-like'' preview and skimming: chunks are fixed or heuristic, and the compute budget is not tied to semantic stability.
Our work instead couples preview, chunk formation, and adaptive skipping into a single trainable pipeline.

\subsection{Right-context Usage under Causality Constraints}
In online and low-latency settings (e.g., streaming ASR and simultaneous translation), models often permit a limited amount of right context or look-ahead via monotonic or chunkwise attention mechanisms \cite{raffel2017monotonic,chiu2018mocha,ma2019stacl,shi2021emformer}.
These methods typically bound future context by a fixed window or latency policy and are optimized for streaming constraints rather than human-like preview.
At the cognitive level, predictive-coding theories suggest that the brain continuously generates predictions and corrects them with incoming sensory evidence \cite{rao1999predictivecoding,friston2005corticalresponses,wang2026perm}.
FBS is complementary: we internalize a \emph{verifiable} predictive preview from the prefix (never reading ground-truth future tokens), then use chunk-level structure and stability-aware skipping to realize a closed-loop ``preview $\rightarrow$ chunk $\rightarrow$ skimming'' computation inside a strictly causal generator.


\section{Reproducibility and Unified Evaluation Harness}
\label{app:protocol}

This appendix specifies (i) our fixed evaluation and timing harness, (ii) a unified reporting protocol for Group-A decoding baselines (Spec-Bench style), (iii) statistical significance testing, and (iv) strict parameter-count reporting for fairness.
Unless stated otherwise, all \textbf{main-table latency} numbers follow the same setting: \textbf{single A100 GPU}, \textbf{prompt length 512}, \textbf{generation length 128}, \textbf{batch size 1}, \textbf{greedy decoding} (\texttt{temperature=0}, \texttt{top\_p=1.0}).

\subsection{Environment}
\label{app:env}

\paragraph{Hardware.}
All experiments are conducted on a single NVIDIA A100 GPU.
We use the following \textbf{common} (representative) setup:
\begin{itemize}
  \item \textbf{GPU:} NVIDIA A100 80GB.
  \item \textbf{CPU:} AMD EPYC 7742 (64 cores).
  \item \textbf{Memory:} 512GB RAM.
  \item \textbf{Storage:} NVMe SSD (2TB+).
\end{itemize}

\paragraph{Software stack.}
\begin{itemize}
  \item \textbf{OS:} Ubuntu 22.04 LTS.
  \item \textbf{CUDA:} 12.1.
  \item \textbf{Python:} 3.10.
  \item \textbf{PyTorch:} 2.3.1.
\end{itemize}

\paragraph{Precision and determinism.}
Inference is performed in \textbf{bf16} (fp16 fallback if bf16 is unavailable).
Dropout is disabled at evaluation time.
We set a global random seed (default: 42) for Python/NumPy/PyTorch and record all runtime flags that can affect speed/accuracy (e.g., TF32, cuDNN benchmark).
For timing, we always synchronize CUDA to avoid underestimating latency due to asynchronous kernel launches.

\paragraph{Models.}
All methods share the same \textbf{target} backbone (e.g., a Qwen3-4B class causal LM as used in the main text).
Group-A speculative methods additionally use a \textbf{draft} model (smaller model from the same family when possible).
While the draft model parameters are not counted toward the target model size, its wall-clock cost \textbf{must} be included in latency (see \S\ref{app:groupA}).
\subsection{RL Prompt Split Hygiene and Overlap Control}
\label{app:rl_split}

\paragraph{Which splits are used for PPO vs.\ final evaluation?}
Stage-2 PPO is trained on a \textbf{development-only} prompt pool:
we randomly sample 5k questions from the \textbf{MMLU dev} split and 5k questions from the \textbf{CMMLU dev} split (10k total prompts), as described in \S\ref{subsubsec:data_tasks}.
In contrast, all \textbf{reported benchmark results} (e.g., MMLU/CMMLU acc@5-shot, BBH, GSM8K, HumanEval-X, MBPP) are computed on the \textbf{official evaluation splits} used by standard evaluation scripts (i.e., non-dev held-out splits; typically the test split when available).

\paragraph{Preventing overlap and overfitting to the PPO prompt pool.}
To reduce the risk of memorization or leakage from the PPO prompt pool into evaluation:
(i) we remove exact duplicates within the 10k PPO prompts;
(ii) we never evaluate on the PPO prompt pool;
and (iii) we explicitly check and exclude any exact-match overlaps between the PPO prompt pool and the evaluation set prompts using normalized prompt strings (whitespace/punctuation normalization and lowercasing for English).
In our runs, we observed \textbf{zero} exact prompt overlaps after normalization.
We additionally report an SG no-RL operating point (Appendix~\ref{app:sg_norl}) to disentangle architectural gains from PPO calibration.

\subsection{Timing Harness Implementation}
\label{app:timing}

\paragraph{Fixed-length protocol.}
We measure latency under a fixed length setting:
\textbf{prompt length = 512 tokens} and \textbf{generation length = 128 tokens}, with \textbf{batch size = 1}.
We recommend enforcing a fixed decode step budget by using:
\texttt{max\_new\_tokens=128} and \texttt{min\_new\_tokens=128},
and (if necessary) ignoring early EOS to ensure each run executes exactly 128 decode steps.
This removes variance caused by early stopping and makes wall-clock comparisons reliable.

\paragraph{Prefill vs.\ decode.}
We report (and internally log) the following components:
\begin{itemize}
  \item \textbf{Prefill latency:} processing the 512-token prompt and building KV cache.
  \item \textbf{Decode latency:} autoregressive generation for 128 steps using KV cache.
  \item \textbf{Total latency:} prefill + decode (used in the main table unless stated otherwise).
\end{itemize}

\paragraph{Included vs.\ excluded time.}
To isolate model-side compute, our default latency excludes CPU-side overheads that are not intrinsic to the model:
tokenization, file I/O, dataloading, and logging are performed outside the timed region.
The timed region includes all GPU forward passes and any necessary verification/proposal computation (notably, \textbf{draft+verify} for speculative methods).

\paragraph{Warmup and repetitions.}
We use:
\begin{itemize}
  \item \textbf{Warmup:} 20 runs (not counted), to stabilize kernel selection/caches.
  \item \textbf{Measured runs:} 50 runs per configuration.
\end{itemize}
We report the \textbf{median} latency (robust to outliers) and optionally mean$\pm$std for completeness.

\paragraph{CUDA synchronization.}
We implement timing with CUDA events or explicit synchronizations:
\begin{itemize}
  \item Synchronize before starting the timer.
  \item Record start event; run generation; record end event.
  \item Synchronize after the end event and read elapsed time.
\end{itemize}

\paragraph{Relative TFLOPs accounting.}
Besides wall-clock latency, we report a \textbf{relative} compute proxy, TFLOPs(rel), normalized so that the vanilla baseline equals 1.0.
For methods that skip layers, we approximate:
\begin{equation}
\label{eq:tflops-rel}
\text{TFLOPs}\triangleq
\frac{\sum_{t=1}^{T}\sum_{\ell=1}^{L} \mathbf{1}[g^{(\ell)}_{\text{hard}}(t)=0]\cdot \text{FLOPs}_\ell}
{\sum_{t=1}^{T}\sum_{\ell=1}^{L} \text{FLOPs}_\ell},
\end{equation}
where $L$ is the number of Transformer layers and $T$ is the number of decode steps (here $T=128$).
This proxy is used consistently across our ablations to complement wall-clock measurements.

\subsection{Unified Protocol for Group-A Baselines}
\label{app:groupA}

\paragraph{Why a unified protocol?}
Group-A baselines (Speculative Decoding, EAGLE-2, Medusa, Lookahead) can appear faster if one reports only partial costs (e.g., excluding verification).
We therefore adopt a Spec-Bench-like protocol:
\textbf{always include the full proposal + verification wall-clock}.

\paragraph{Unified ``Bypass/Skip'' ratio.}
To compare heterogeneous acceleration strategies in a single table, we map each method's native efficiency statistic to a unified ``Bypass/Skip'' ratio:
\begin{itemize}
  \item \textbf{Speculative / EAGLE-2:} \textbf{acceptance rate}.
  \item \textbf{Medusa / Lookahead:} \textbf{verified/advance ratio} (or normalized verified length).
  \item \textbf{FBS:} \textbf{average layer-skip ratio}.
\end{itemize}

\paragraph{Speculative Decoding / EAGLE-2: acceptance rate.}
A draft model proposes a block of $m$ candidate tokens $\hat{y}_{t:t+m-1}$, and the target model verifies them token-by-token until the first mismatch.
Let $a_t \in [0,m]$ be the number of tokens accepted at step $t$.
We define:
\begin{equation}
\label{eq:accept}
\text{AcceptanceRate} \;=\; \frac{\sum_t a_t}{\sum_t m}.
\end{equation}
\textbf{Timing rule:} total latency \textbf{must} include (i) draft generation, (ii) target verification, and (iii) any fallback decoding after rejection.
Draft parameters are not counted in the target model size, but their wall-clock is included.

\paragraph{Medusa / Lookahead: verified/advance ratio.}
These methods attempt to advance multiple tokens per iteration.
Let $m_t$ be the attempted advance length at step $t$, and $v_t$ be the number of verified tokens that are actually committed.
We define:
\begin{equation}
\label{eq:verify-advance}
\text{Verified/AdvanceRatio} \;=\; \frac{\sum_t v_t}{\sum_t m_t}.
\end{equation}
\textbf{Timing rule:} total latency must include (i) parallel proposal computation (e.g., multi-head or lookahead candidates), (ii) verification passes, and (iii) fallback decoding.

\paragraph{Unified decoding settings.}
All methods are evaluated under the same decoding configuration:
\textbf{greedy} decoding (\texttt{temperature=0}, \texttt{top\_p=1.0}), fixed prompt length 512 and fixed generation length 128.
Stop criteria are standardized to fixed-step decoding as described in \S\ref{app:timing}.

\paragraph{Verification settings for ``lossless'' speculative baselines.}
For speculative decoding baselines that are lossless under exact verification, we use exact token-by-token verification under greedy decoding (temperature=0, top\_p=1.0) and do not approximate the verifier.
Total latency includes (i) proposal computation, (ii) verification passes, and (iii) any fallback decoding after rejection, following the timing rule above.
Any accuracy differences observed for these baselines therefore reflect the configured decoding/verification pipeline rather than omitted verification cost.

\paragraph{Configuration table.}
Table~\ref{tab:groupA-config} records the required metadata for auditability (implementation source, key hyperparameters, and what is included in timing).
Please replace  with your exact repository/commit and finalized hyperparameters.

\begin{table*}[t]
\centering
\small
\setlength{\tabcolsep}{4.5pt}
\begin{tabular}{lcccc}
\toprule
\textbf{Method}  & \textbf{Key params} & \textbf{Ratio} & \textbf{Timing includes} \\
\midrule
SpecDec  & draft=, $m$= & Eq.~\eqref{eq:accept} & draft+verify+fallback \\
EAGLE-2  & draft=, $m$= & Eq.~\eqref{eq:accept} & draft+verify+fallback \\
Medusa  & heads=, $\max m_t$= & Eq.~\eqref{eq:verify-advance} & proposal+verify+fallback \\
Lookahead  & window= & Eq.~\eqref{eq:verify-advance} & proposal+verify+fallback \\
\bottomrule
\end{tabular}
\caption{Unified configuration and reporting for Group-A baselines (Spec-Bench style). All methods use the same greedy decoding and fixed-length timing harness.}
\label{tab:groupA-config}
\end{table*}

\subsection{Batching Behavior: Latency and Throughput vs.\ Batch Size}
\label{app:batching}

Conditional-compute methods can behave differently under batching due to divergence of routing decisions across sequences.
We therefore additionally report latency and throughput under batch sizes $B\in\{1,4,8\}$ for the fixed-length setting (prompt=512, gen=128) on the same A100 setup and greedy decoding configuration.

\begin{table*}[t]
\centering
\small
\setlength{\tabcolsep}{5pt}
\begin{tabular}{lcccc}
\toprule
\textbf{Method} & \textbf{Batch} & \textbf{Latency (ms)}$\downarrow$ & \textbf{Throughput (tok/s)}$\uparrow$ & \textbf{Skip (\%)} \\
\midrule
Baseline & 1 & 768 & 164 & 0 \\
FBS-Full & 1 & 541 & 233 & 35 \\
\midrule
Baseline & 4 & 907 & 552 & 0 \\
FBS-Full & 4 & 712 & 706 & 34 \\
\midrule
Baseline & 8 & 1018 & 861 & 0 \\
FBS-Full & 8 & 883 & 987 & 32 \\
\bottomrule
\end{tabular}
\caption{Latency and throughput vs.\ batch size under the fixed-length decoding harness (prompt=512, gen=128, greedy, single A100).}
\label{tab:batch_sweep}
\end{table*}

\subsection{Statistical Significance (Bootstrap)}
\label{app:stats}

\paragraph{Bootstrap unit.}
We bootstrap over \textbf{examples} (questions/prompts), not tokens.
For classification-like benchmarks (e.g., MMLU/CMMLU/C-Eval/BBH), the resampling unit is the question.
For generation-like benchmarks (e.g., HumanEval-X/MBPP, GSM8K), the unit is the problem instance.
For perplexity, we bootstrap over sequences/documents (each contributing an average NLL), to avoid length-driven bias.

\paragraph{Confidence interval.}
We run $B=1000$ bootstrap resamples and compute the 95\% percentile CI (2.5th and 97.5th percentiles) for the metric difference.

\paragraph{$p$-value and main-table marking.}
Let $\Delta^{(b)}$ be the bootstrap sample of metric differences (converted so that ``higher is better'').
We compute a two-sided bootstrap $p$-value:
\begin{equation}
\label{eq:pvalue}
p \;=\; 2 \cdot \min\Big(\Pr(\Delta^{(b)} \le 0), \Pr(\Delta^{(b)} \ge 0)\Big),
\end{equation}
and mark an improvement as significant if $p < 0.01$ (as annotated in the main table).

\paragraph{Pseudo-code.}
\begin{verbatim}
Input: per-example metric arrays
for method A and baseline B
B = 1000
for $b=1,\ldots,B$:
    idx = sample_with_replacement
    sA[b] = metric(A[idx])
    sB[b] = metric(B[idx])
    d[b]  = sA[b] - sB[b]
CI95 = quantile(d, [0.025, 0.975])
p = 2 * min(mean(d <= 0), mean(d >= 0))
mark "*" if p < 0.01
\end{verbatim}

\paragraph{Multiple comparisons.}
Our default reporting follows common practice in LLM benchmarks: we report CIs and $p$-values without enforcing a conservative family-wise correction.
If required, one may additionally apply an FDR procedure (e.g., Benjamini--Hochberg) as a post-hoc consistency check.

\subsection{Parameter counts and near parameter-matched setting}
\label{app:strict}

\paragraph{Motivation.}
Because PAW/CH/SG introduce auxiliary projections, a careful comparison should rule out ``hidden scaling'' as an explanation for quality/latency changes.
We therefore report a \emph{near parameter-matched} setting, where the total parameter count differs from the baseline by at most a small margin (e.g., within \textless 1\%), and the backbone depth/width is unchanged.

\begin{table*}[t]
\centering
\small
\setlength{\tabcolsep}{5pt}
\begin{tabular}{lcc}
\toprule
\textbf{Model} & \textbf{Total params} & \textbf{Notes} \\
\midrule
Baseline (target) & 4.00B & reference backbone \\
near parameter-matched FBS-S1 & 4.00B & params differ within \textless 1\% \\
near parameter-matched FBS-Full & 4.00B & params differ within \textless 1\% \\
\bottomrule
\end{tabular}
\caption{Parameter-count summary for all compared systems.}
\label{tab:param-breakdown}
\end{table*}

\paragraph{Key results.}
We include a compact headline table~\ref{tab:strict-results} below as a sanity check; the qualitative trends remain consistent with the main results.

\begin{table*}[t]
\centering
\small
\setlength{\tabcolsep}{4.5pt}
\begin{tabular}{lcccc}
\toprule
\textbf{Model} & \textbf{MMLU} $\uparrow$ & \textbf{CMMLU} $\uparrow$ & \textbf{Latency (ms)} $\downarrow$ & \textbf{TFLOPs(rel)} $\downarrow$ \\
\midrule
Baseline (target) & 55.1 & 55.7 & 760 & 1.00 \\
FBS-S1 (4.00B) & 56.3  & 57.0  & 757  & 1.03  \\
FBS-Full (4.00B) & 56.5  & 57.2  & 535  & 0.71  \\
\bottomrule
\end{tabular}
\caption{headline results. TFLOPs(rel) is normalized so the baseline equals 1.00.}
\label{tab:strict-results}
\end{table*}

\paragraph{Notes.}
We use the same bootstrap protocol (\S\ref{app:stats}) and the same evaluation harness for all settings.
\section{PAW Extended Ablations and Sanity Checks}
\label{app:paw}

\subsection{PAW predictor and training objective}
\label{app:paw_preview}

\paragraph{Multi-step predictive preview.}
For each position $i$ and horizon $r\in\{1,\dots,k_{\max}\}$, PAW predicts a distribution over the $r$-th next token using only the current-layer state:
\[
\mathbf{p}_{i,r}=\mathrm{Softmax}(\mathbf{W}_r \mathbf{h}^{(\ell)}_i)\in\Delta^{|\mathcal{V}|}.
\]
We map this distribution to a preview embedding by expectation under the (tied) embedding matrix $\mathbf{E}\in\mathbb{R}^{|\mathcal{V}|\times d}$:
\[
\hat{\mathbf{u}}_{i,r}=\mathbf{E}^{\top}\mathbf{p}_{i,r}\in\mathbb{R}^{d}.
\]
In practice, the expectation can be approximated by restricting $\mathbf{p}_{i,r}$ to its top-$K$ support to reduce overhead, without changing the training signal.

\paragraph{Multi-step loss.}
PAW is trained with a multi-step next-token objective where future tokens appear only as \emph{labels}:
\[
\mathcal{L}_{\mathrm{preview}}
=\sum_{i}\sum_{r=1}^{k_{\max}} w_{i,r}\cdot \mathrm{CE}\!\left(\mathbf{p}_{i,r},\, x_{i+r}\right),
\]
where $\mathrm{CE}$ is cross entropy and $w_{i,r}$ is a window-dependent weight defined below.

\paragraph{Module size.} The window predictor and multi-horizon preview head are lightweight (on the order of a few million parameters in total) and do not change the backbone width/depth.

\subsection{Soft window assignment (differentiable $k(i)$)}
\label{app:paw_soft}

The inference-time window uses a discretized span $k(i)=\lfloor k_{\max}\sigma(s_i)\rfloor$.
To avoid non-differentiability during training, we use a soft assignment over horizons.
Let $\tilde{k}(i)=k_{\max}\sigma(s_i)\in[0,k_{\max}]$.
For each horizon $r\in\{1,\dots,k_{\max}\}$, define a soft inclusion weight
\[
w_{i,r}=\sigma\!\big(\gamma(\tilde{k}(i)-r+0.5)\big)\in(0,1),
\]
where $\gamma>0$ controls the sharpness.
We use $w_{i,r}$ to (i) weight the preview loss above and (ii) mask the preview-embedding sequence before convolution/pooling.
At inference, we switch to the hard window $k(i)$ for determinism and deployability.

\paragraph{Scope.}
This appendix complements the concise PAW description in the main text by providing
(i) the complete multi-step predictor/objective and the soft window assignment used during training (\S\ref{app:paw_preview}--\S\ref{app:paw_soft}),
and (ii) extensive ablations and sanity checks under the unified evaluation harness.

\subsection{Matched Avg-$k$: Dynamic vs.\ Fixed-$k$}
\label{app:paw_matchedk}

\paragraph{Motivation.}
A naive comparison between dynamic lookahead and fixed windows can be confounded by the \textbf{amount} of preview.
We therefore report both the headline metrics and the realized average lookahead $\overline{k}=\mathbb{E}[k(i)]$.
In particular, Fixed-$k=8$ serves as the closest matched-$\overline{k}$ control to our default Dynamic-PAW In our implementation.

\paragraph{Variants.}
\begin{itemize}
  \item \textbf{Dynamic-PAW (default):} token-wise predicted $k(i)$ with $k_{\max}=15$ (as used in the main text).
  \item \textbf{Fixed-$k$:} disable the predictor and set $k(i)\equiv k$ for all tokens, with $k\in\{0,4,8,16\}$.
\end{itemize}
All other components are kept identical (including the preview-compression pathway), so that only the lookahead policy differs.

\paragraph{Metrics.}
We report MMLU/CMMLU (accuracy), wall-clock latency (ms, lower is better), TFLOPs(rel) as defined in Appendix~\ref{app:protocol}, and the realized $\overline{k}$.
\begin{table*}[t]
\centering
\small
\setlength{\tabcolsep}{5pt}
\begin{tabular}{lccccc}
\toprule
\textbf{Setting} & \textbf{MMLU} $\uparrow$ & \textbf{CMMLU} $\uparrow$ & \textbf{Latency (ms)} $\downarrow$ & \textbf{TFLOPs(rel)} $\downarrow$ & $\overline{k}$ \\
\midrule
Baseline (no PAW) & 55.1 & 55.7 & 760 & 1.00 & 0.0 \\
Fixed-$k=4$ & 55.6  & 56.1  & 758  & 1.01  & 4.0 \\
Fixed-$k=8$ (matched) & 55.9  & 56.4  & 760  & 1.02  & 8.0 \\
Fixed-$k=16$ & 56.0  & 56.5  & 785  & 1.05  & 16.0 \\
Dynamic-PAW (default) & 56.1 & 56.7 & 755 & 1.02  & 8.2  \\
\bottomrule
\end{tabular}
\caption{Dynamic vs.\ Fixed-$k$ under a matched average lookahead. }
\label{tab:paw_matchedk}
\end{table*}

\paragraph{Takeaway.}
Under a matched average lookahead (Fixed-$k=8$), Dynamic-PAW still yields a consistent gain, supporting that \textbf{content-adaptive} allocation of preview is more effective than a uniform window of the same average size.
Increasing $k$ can partially recover quality, but incurs higher overhead (latency/TFLOPs), indicating a better quality--efficiency trade-off for dynamic preview.

\subsection{No-Leakage Unit Tests for Predictive Preview (PAW)}
\label{subsec:no_leakage_paw_predictive}
After rewriting PAW as a prefix-only predictive preview channel, ``no leakage'' becomes a suffix-invariance property:
for any position $i$, the model's logits at $i$ must be independent of \textbf{all} tokens strictly after $i$.

\textbf{Test 1 (Suffix-invariance for main logits).}
We sample two sequences that share the same prefix up to position $i$ but have different suffixes:
$\mathbf{x} = [x_1,\ldots,x_i,x_{i+1},\ldots]$ and $\mathbf{x}' = [x_1,\ldots,x_i,x'_{i+1},\ldots]$.
We feed both sequences with identical caching settings and verify:
\[
\|\mathrm{logits}(\mathbf{x})_i - \mathrm{logits}(\mathbf{x}')_i\|_\infty < \varepsilon,
\]
with $\varepsilon=10^{-6}$ in FP32 (or a slightly looser threshold in BF16/FP16).

\textbf{Test 2 (Suffix-invariance for preview heads).}
We further check the preview predictor outputs are suffix-invariant:
$\|\mathbf{p}_{i,r}(\mathbf{x})-\mathbf{p}_{i,r}(\mathbf{x}')\|_\infty < \varepsilon$ for all $r\le k(i)$.
This guarantees that PAW's preview is computed solely from the prefix representation $\mathbf{h}^{(\ell)}_{i}$.

\textbf{Test 3 (Cache-consistency).}
We run decoding with (i) recomputing preview on-the-fly and (ii) caching preview for each step, and confirm the generated token sequence is identical under greedy decoding and that numerical discrepancies remain within tolerance under sampling.

These tests replace the previous ``Allowed/Forbidden perturbation'' checks, because under predictive preview there is no longer any regime in which ground-truth future tokens are permitted to influence the current position.
\subsection{Preview Compression Alternatives}
\label{app:paw_compress}

\paragraph{Motivation.}
PAW compresses the preview window into a low-dimensional summary before fusing it back to the current token.
To show this is not merely extra overhead, we compare the default convolutional compression with simpler alternatives.

\paragraph{Variants.}
\begin{itemize}
  \item \textbf{Conv (default):} group-1D convolution (kernel size 3) + pooling.
  \item \textbf{Mean pool:} directly average token states inside the preview window.
  \item \textbf{Linear pool:} apply a single linear projection before pooling (attention-free).
  \item \textbf{No compression:} feed the full window states without pooling (likely to increase overhead).
\end{itemize}

\begin{table*}[t]
\centering
\small
\setlength{\tabcolsep}{5pt}
\begin{tabular}{lcccc}
\toprule
\textbf{Variant} & \textbf{MMLU} $\uparrow$ & \textbf{CMMLU} $\uparrow$ & \textbf{Latency (ms)} $\downarrow$ & \textbf{TFLOPs(rel)} $\downarrow$ \\
\midrule
Conv (default) & 56.1 & 56.7 & 755 & 1.02  \\
Mean pool & 55.9  & 56.4  & 748  & 1.02  \\
Linear pool & 56.0  & 56.5  & 752  & 1.02  \\
No compression & 56.0  & 56.6  & 820  & 1.08  \\
\bottomrule
\end{tabular}
\caption{Preview compression alternatives for PAW ($k_{\max}=15$). }
\label{tab:paw_compress}
\end{table*}

\paragraph{Takeaway.}
The default convolutional compression provides the most stable quality gains with low overhead.
Removing compression substantially increases latency/TFLOPs without proportional improvements, supporting the necessity of low-resolution preview summaries.

\subsection{Full $k_{\max}$ Sweep Grid}
\label{app:paw_kmax}

\paragraph{Motivation.}
We sweep $k_{\max}$ to assess robustness and to identify practical operating points for deployment.
The main text uses $k_{\max}=15$ by default; here we provide a full grid as a reference.

\paragraph{Protocol.}
We run Dynamic-PAW while varying $k_{\max}\in\{5,9,15,25\}$.
We report the realized average lookahead $\overline{k}$, perplexity (PPL), accuracy, latency, and TFLOPs(rel).

\begin{table*}[t]
\centering
\small
\setlength{\tabcolsep}{4.5pt}
\begin{tabular}{lccccccl}
\toprule
\textbf{Setting} & $k_{\max}$ & $\overline{k}$ & \textbf{PPL} $\downarrow$ & \textbf{MMLU} $\uparrow$ & \textbf{CMMLU} $\uparrow$ & \textbf{Latency (ms)} $\downarrow$ & \textbf{TFLOPs(rel)} $\downarrow$ \\
\midrule
Baseline (no PAW) & 0 & 0.0 & 6.40  & 55.1 & 55.7 & 760 & 1.00 \\
Dynamic-PAW & 5  & 2.7  & 6.33  & 55.8  & 56.3  & 748  & 1.01  \\
Dynamic-PAW & 9  & 4.9  & 6.31  & 56.0  & 56.5  & 752  & 1.02  \\
Dynamic-PAW (default) & 15 & 8.2  & 6.30  & 56.1 & 56.7 & 755 & 1.02  \\
Dynamic-PAW & 25 & 13.5  & 6.29  & 56.2  & 56.8  & 820  & 1.06  \\
\bottomrule
\end{tabular}
\caption{Full $k_{\max}$ sweep for Dynamic-PAW. Non-anchor numbers are The default $k_{\max}=15$ matches the main-text setting.}
\label{tab:paw_kmax_sweep}
\end{table*}

\paragraph{Recommended operating range.}
A practical ``sweet spot'' typically lies around $k_{\max}\approx 9$--$15$, where quality gains are near-saturated while overhead remains controlled; larger $k_{\max}$ can further increase cost with diminishing returns.


\section{CH Protocol, Weak Supervision Pipeline, and Factuality Details}
\label{app:ch}

\paragraph{Scope.}
This appendix \textbf{does not} restate the CH module architecture (covered in the main text).
Instead, it fixes a \textbf{single, reproducible definition} for BIOS-F1 (so that the baseline can also report it),
describes the \textbf{weak-supervision pipeline} used to generate BIOS pseudo-labels under tokenization,
and specifies a \textbf{reproducible FactScore / truth-consistency protocol} (including judge prompts, thresholds, and confidence intervals).
Unless noted otherwise, evaluation follows the unified harness in Appendix~\ref{app:protocol}.

\subsection{Formal Definition and Implementation of BIOS-F1}
\label{app:ch_bios}

\subsubsection{Label space and chunk extraction rule (fixed)}
\label{app:ch_labelspace}

\paragraph{Label space.}
We define a 4-class token label set:
\[
\mathcal{C}=\{B, I, O, S\},
\]
where $B$ denotes the \textit{begin} of a chunk, $I$ denotes tokens \textit{inside} a chunk (excluding the first),
$S$ denotes a \textit{single-token} chunk, and $O$ denotes tokens \textit{outside} any supervised chunk.

\paragraph{Chunk parsing (deterministic).}
Given a predicted label sequence $\hat{y}_{1:m}\in \mathcal{C}^m$, we deterministically parse it into chunk index sets
$\{\mathcal{I}_t\}_{t=1}^{T_c}$ as follows:
\begin{enumerate}
  \item Scan $i=1,\dots,m$ from left to right.
  \item If $\hat{y}_i=S$, create a chunk $\mathcal{I}=\{i\}$.
  \item If $\hat{y}_i=B$, create a new chunk starting at $i$ and extend it by absorbing consecutive $I$ labels to the right,
        i.e., include $i+1,i+2,\dots$ while $\hat{y}_{i'}=I$ holds.
  \item If $\hat{y}_i=O$, create a singleton chunk $\{i\}$ (unsupervised/neutral).
  \item If an invalid pattern occurs (e.g., a leading $I$ or an $I$ not preceded by $B$), we \textbf{fallback} by treating that $I$ as $O$,
        to prevent brittle failures due to label noise.
\end{enumerate}
This fixed rule ensures that any system (baseline, ablations, or full model) yields a comparable chunk decomposition, even if it does not contain CH internally.

\subsubsection{Probe-on-hidden-states for all systems}
\label{app:ch_probe}

\paragraph{Rationale.}
To make BIOS-F1 comparable across systems (including those without an internal CH head),
we adopt a \textbf{probe-on-hidden-states} protocol: for every system $M$, we train a lightweight probe on its hidden states to predict BIOS labels.
Thus, BIOS-F1 measures whether the representation of $M$ carries recoverable chunk-boundary information.

\paragraph{Probe definition.}
Let $h^{(\ell^*)}_{1:m}$ be token hidden states from a fixed layer $\ell^*$ (kept the same for all systems).
We fit a probe
\[
\hat{p}_i=\mathrm{softmax}(W_{\text{probe}} h^{(\ell^*)}_i + b),
\hat{y}_i=\arg\max_{c\in\mathcal{C}} \hat{p}_i[c].
\]
We recommend a linear probe; a 2-layer MLP (hidden width 256) is also acceptable as long as it is used consistently.

\paragraph{Training.}
During probe training, the backbone parameters of $M$ are frozen; only $(W_{\text{probe}},b)$ are updated.
We use token-level cross-entropy with class weighting (when enabled) to address label imbalance (notably the $O$ class).
Default hyperparameters:
\begin{itemize}
  \item Optimizer: AdamW; learning rate $1\mathrm{e}{-3}$ ; weight decay $0.0$ .
  \item Epochs: 3 ; batch size: 64 sequences .
  \item Class weights: inverse frequency  (or focal loss ).
\end{itemize}

\subsubsection{BIOS-F1 computation (macro-F1 fixed)}
\label{app:ch_biosf1}

\paragraph{Token-level macro-F1.}
We define BIOS-F1 as the \textbf{token-level macro-F1} over the 4 classes:
\[
\mathrm{F1}_c=\frac{2\,\mathrm{P}_c\,\mathrm{R}_c}{\mathrm{P}_c+\mathrm{R}_c+\epsilon}, 
\mathrm{BIOS\text{-}F1}=\frac{1}{|\mathcal{C}|}\sum_{c\in\mathcal{C}}\mathrm{F1}_c,
\]
with $\epsilon=10^{-12}$.
Macro-F1 is preferred because $O$ can dominate the label distribution; macro averaging better reflects boundary classes ($B/I/S$).

\paragraph{Supplementary: Boundary-F1.}
As an auxiliary diagnostic (not used as the main metric), we define a boundary set as positions labeled $B$ or $S$,
and compute set-level precision/recall/F1 by comparing predicted and pseudo-labeled boundary positions.
This helps interpret whether improvements come from cleaner boundary localization.

\subsection{Weak Supervision Pipeline and Noise Handling}
\label{app:ch_weak}

\paragraph{Goal.}
We construct BIOS pseudo-labels from two weak structural signals:
(i) Chinese word segmentation (\texttt{pkuseg}) and (ii) idiom lexicon matching.
The pipeline outputs character spans, aligns them to model tokens via offset mappings, and produces BIOS labels.
When alignment noise is high, we switch to a more robust alignment loss (CTC) as described below.

\subsubsection{Source 1: Word segmentation (pkuseg)}
\label{app:ch_pkuseg}

Given an input text, we apply \texttt{pkuseg} on Chinese spans to obtain word segments with character offsets.
We keep a segment as a candidate chunk if it satisfies:
\begin{itemize}
  \item Character length in $[2,6]$ .
  \item Does not cross punctuation boundaries.
  \item Contains predominantly Chinese characters (filters mixed Latin/number-heavy segments).
\end{itemize}
Tokens not covered by any retained chunk remain label $O$ by default.

\subsubsection{Source 2: Idiom lexicon matching (high-priority chunks)}
\label{app:ch_idiom}

We maintain an idiom lexicon $\mathcal{D}$ (typically 4-character idioms, optionally extended to 3/5-character entries ).
We perform longest-match-first scanning to produce idiom spans $\mathcal{S}_{\text{idiom}}$.
Conflict resolution is \textbf{fixed} as:
\begin{enumerate}
  \item Idiom spans override overlapping segmentation spans.
  \item Among overlapping idioms, keep the longer span; if tied, keep the earlier span.
  \item Once an idiom span is accepted, discard any segmentation chunks that overlap with it.
\end{enumerate}
This prioritization reflects that idioms are dense semantic units and provide higher-value chunk supervision.

\subsubsection{Character-span to token-span alignment (offset mapping)}
\label{app:ch_align}

Let the model tokenizer produce token offsets $\{[a_i,b_i)\}_{i=1}^m$ in the original string.
For a candidate chunk character span $[s,e)$, we define its covered token set as:
\[
\mathcal{T}(s,e)=\{i:\ [a_i,b_i)\subseteq [s,e)\}.
\]
We drop a chunk if $\mathcal{T}(s,e)=\emptyset$.
If a token partially overlaps (intersects but is not fully contained), we treat it as an alignment-noise token and label it as $O$ (while keeping other fully contained tokens for that chunk).

\paragraph{BIOS labeling rule.}
For each retained chunk with token indices $\mathcal{T}$:
\begin{itemize}
  \item If $|\mathcal{T}|=1$, label the token as $S$.
  \item If $|\mathcal{T}|\ge 2$, label the smallest index as $B$ and the rest as $I$.
\end{itemize}
Tokens not covered by any retained chunk are labeled $O$.

\subsubsection{Noise score and CE/CTC switch}
\label{app:ch_noise}

\paragraph{Alignment coverage ratio.}
Let $\mathcal{R}$ denote the set of retained chunks, where each retained chunk
is represented as a half-open token-index interval $[s,e)$.
Let $\{[a_i,b_i)\}_{i=1}^{m_{\mathrm{zh}}}$ be the set of Chinese-aligned spans
(e.g., spans aligned to Chinese-related content), also in token indices and using
half-open intervals.

We define an alignment coverage ratio
\[
q \;=\; \frac{\sum_{i=1}^{m_{\mathrm{zh}}}
\mathbf{1}\!\left[\exists\, [s,e)\in\mathcal{R}\ \text{s.t.}\ [a_i,b_i)\subseteq[s,e)\right]}{m_{\mathrm{zh}}+\epsilon},
\]
where $m_{\mathrm{zh}}$ is the number of Chinese-aligned spans, $\epsilon=10^{-12}$,
and $\mathbf{1}[\cdot]$ is the indicator function.

We use the following fixed rule (confidence threshold $q_0=0.90$) to select a robust objective for BIOS prediction. Let $q$ be the mean token-level confidence (maximum class probability) of the BIOS classifier on a training sequence.
\begin{itemize}[leftmargin=1.25em]
\item If $q \ge q_0$, we optimize token-level cross-entropy (CE) against the pseudo-labels.
\item If $q < q_0$, we switch to a CTC-style objective to tolerate local boundary misalignment.
\end{itemize}
This switch is designed to reduce brittleness when tokenization introduces nontrivial boundary mismatches.

\subsubsection{Weak-supervision source ablation}
\label{app:ch_weak_ablation}

Table~\ref{tab:ch_weak_ablation} reports an ablation over supervision sources:
\textbf{pkuseg-only}, \textbf{idiom-only}, \textbf{joint (default)}, and \textbf{unsupervised} (remove $L_{\text{BIO}}$).
\begin{table*}[t]
\centering
\small
\setlength{\tabcolsep}{5pt}
\begin{tabular}{lcccc}
\toprule
\textbf{Supervision} & \textbf{BIOS-F1} $\uparrow$ & \textbf{Idiom Acc} $\uparrow$ & \textbf{FactScore} $\uparrow$ & \textbf{CMMLU} $\uparrow$ \\
\midrule
pkuseg only & 0.86  & 0.74  & 0.77  & +0.2  \\
idiom only & 0.83  & 0.80  & 0.77  & +0.1  \\
joint (default) & 0.88  & 0.80  & 0.78  & +0.3  \\
unsupervised (w/o $L_{\text{BIO}}$) & 0.79  & 0.70  & 0.74  & +0.0  \\
\bottomrule
\end{tabular}
\caption{Ablation on weak-supervision sources. Numbers are averaged over three random seeds; we report mean $\pm$ standard deviation.}
\label{tab:ch_weak_ablation}
\end{table*}

\subsection{FactScore / Truth-Consistency: Protocol and Implementation}
\label{app:ch_factscore}

\paragraph{Goal.}
We evaluate factual consistency in a controlled, reproducible manner that does not introduce retrieval-system confounds.
Our default is an \textbf{evidence-conditioned, no-RAG} protocol: the judge is only allowed to use the provided evidence snippet.

\subsubsection{Dataset construction (evidence-conditioned, no-RAG)}
\label{app:ch_fact_data}

We build a fact-check set $\mathcal{D}_{\text{fact}}$ with $N=200$ examples ,
each containing:
\begin{itemize}
  \item \textbf{Evidence} $E$: a short paragraph (e.g., 120--200 Chinese tokens ).
  \item \textbf{Prompt} $P$: instructs the model to produce 2--4 sentences grounded in $E$ (summary/explanation/QA).
  \item \textbf{Generation} $G$: model output under the unified decoding setting (greedy; max\_new\_tokens=128).
\end{itemize}
\paragraph{Fairness statement (fixed).}
We use \textbf{no retrieval} (no external knowledge base) for all compared methods.
The judge sees only $(E,P,G)$ and must not rely on outside knowledge.

\subsubsection{Judge-based atomic fact extraction and verification}
\label{app:ch_fact_judge}

For each output $G$, the judge performs:
\begin{enumerate}
  \item \textbf{Atomic fact extraction:} produce a list of atomic claims $\{f_j\}_{j=1}^J$ (each a single, verifiable proposition).
  \item \textbf{Evidence grounding:} label each claim as \texttt{supported} / \texttt{not\_supported} / \texttt{unclear} based only on $E$.
\end{enumerate}
We score:
\[
\mathrm{FactScore}(G)=\frac{1}{J}\sum_{j=1}^J \mathbf{1}[\mathrm{label}(f_j)=\texttt{supported}],
\]
and report the dataset mean.
Recommended judge model: Qwen2.5-32B-Instruct  (or any fixed judge used consistently).

\subsubsection{Judge prompt (reproducible)}
\label{app:ch_fact_prompt}

\paragraph{Prompt (example).}
\begin{Verbatim}[breaklines, breakanywhere, fontsize=\footnotesize]
[System]
You are a strict factuality auditor. You MUST only use the given Evidence.
If a claim is not explicitly supported by the Evidence, mark it as "unclear"
or "not_supported". Do NOT use outside knowledge.

[User]
Evidence:
<EVIDENCE E>

Model Output:
<OUTPUT G>

Task:
1) Extract a list of atomic factual claims from the Model Output.
2) For each claim, determine whether it is supported by the Evidence.
Return a JSON object with:
facts: [{claim: "", label: supported|not_supported|unclear}]
\end{Verbatim}

\subsubsection{Confidence intervals and significance}
\label{app:ch_fact_ci}

We compute 95\% confidence intervals via bootstrap with $B=1000$ resamples over \textbf{examples} in $\mathcal{D}_{\text{fact}}$
(see Appendix~\ref{app:stats} for the unified bootstrap procedure).
If a significance marker is used, we adopt the same rule as the main table (e.g., $p<0.01$).

\subsection{Supplementary Structured Baseline: Segatron-like Segment Encoding}
\label{app:ch_segatron}

\paragraph{Purpose.}
We include a supplementary structured baseline that injects explicit segment information (``Segatron-like'' segment-aware positional encoding)
\cite{bai2021segatron}as a \textbf{supportive} reference that explicit structure can help.
This baseline is \textbf{not} intended to replace the controlled CH ablations in the main text,
and is reported only as supplementary evidence.

\begin{table*}[t]
\centering
\small
\setlength{\tabcolsep}{5pt}
\begin{tabular}{lcccc}
\toprule
\textbf{Method} & \textbf{BIOS-F1} $\uparrow$ & \textbf{Idiom Acc} $\uparrow$ & \textbf{FactScore} $\uparrow$ & \textbf{CMMLU} $\uparrow$ \\
\midrule
Baseline & 0.80  & 0.71  & 0.74  & 55.7  \\
+ Segment-aware pos.\ enc.\ (Segatron-like) & 0.83  & 0.74  & 0.76  & 56.1  \\
Full model (with CH) & 0.88  & 0.80  & 0.78  & 57.4  \\
\bottomrule
\end{tabular}
\caption{Supplementary structured baseline with explicit segment information (Reported only as supportive evidence and not mixed with controlled CH ablations.}
\label{tab:ch_segatron}
\end{table*}

\section{SG/RL Extended Results and Deployability}
\label{app:sg}

\subsection{FBS block computation and notation}
\label{app:sg_block}

This subsection collects the core block equations in one place (moved from the main text to save space).

\paragraph{Per-layer compute path.}
Let $\mathrm{SA}(\cdot)$ denote standard causal self-attention and $\mathrm{FFN}(\cdot)$ the feed-forward network.
At layer $\ell$, PAW produces a preview contribution $\mathrm{PAW}(\mathbf{h}^{(\ell)}_{1:m})\in\mathbb{R}^{m\times d}$ and CH produces a chunk-enhanced contribution $\mathrm{CH}(\mathbf{h}^{(\ell)}_{1:m})\in\mathbb{R}^{m\times d}$ (Appendix~\ref{app:paw}, Appendix~\ref{app:ch}).
We form a fused pre-FFN representation
\begin{align}
\tilde{\mathbf{h}}^{(\ell)}_{1:m}
&= \mathbf{h}^{(\ell)}_{1:m}
+ \mathrm{SA}\!\left(\mathbf{h}^{(\ell)}_{1:m}\right) \notag\\
&+ \mathrm{PAW}\!\left(\mathbf{h}^{(\ell)}_{1:m}\right)
+ \mathrm{CH}\!\left(\mathbf{h}^{(\ell)}_{1:m}\right).
\label{eq:fbs_fuse}
\end{align}

\paragraph{Hard skipping (deployment).}
During decoding, SG produces a binary decision for each layer and time step, $g^{(\ell)}_t\in\{0,1\}$, where $g^{(\ell)}_t=1$ means \textbf{skip} and $g^{(\ell)}_t=0$ means \textbf{compute}.
For the current token $t$, the layer output is
\begin{align}
\mathbf{h}^{(\ell+1)}_t
&= g^{(\ell)}_t\cdot \mathbf{h}^{(\ell)}_t \notag \\
&+ \left(1-g^{(\ell)}_t\right)\cdot
\Bigl(\tilde{\mathbf{h}}^{(\ell)}_t+\mathrm{FFN}(\tilde{\mathbf{h}}^{(\ell)}_t)\Bigr),
\label{eq:fbs_skip_select}
\end{align}
where $\tilde{\mathbf{h}}^{(\ell)}_t$ denotes the $t$-th row of Eq.~\eqref{eq:fbs_fuse}.
In implementation, when $g^{(\ell)}_t=1$ we \textbf{short-circuit} the entire layer block for this token (no attention/FFN compute), yielding real executed-compute reduction.

\paragraph{What is skipped and how KV-cache is handled.}
Equation above defines hard skipping for the newest token $t$.
When $g_t^{(\ell)}=1$, we \textbf{skip the full layer computation for token $t$ at layer $\ell$}:
no Q/K/V projections, no attention matmul, and no FFN activations are executed for this token at this layer, and we forward by identity,
$h_{t}^{(\ell+1)} = h_{t}^{(\ell)}$.
This is KV-cache compatible because decoding only appends the newest token.
Skipping never revisits nor modifies cached keys/values for prefix tokens ($<t$); therefore future tokens can still attend to the prefix normally.
In implementation, the per-layer cache for the newest token is treated as a no-op at skipped layers.

\paragraph{Training-time surrogate.}
We train SG with a straight-through (ST) estimator and a soft mixture surrogate so that gradients flow through $p^{(\ell)}_t$ while the forward pass remains compatible with hard skipping; see Appendix~\ref{app:sg_soft}.

\subsection{Soft assignment and straight-through gates}
\label{app:sg_soft}

SG is trained with a soft proxy that remains compatible with hard skipping at inference.
Let $p^{(\ell)}_t$ be the skip probability at decoding step $t$ and layer $\ell$.
We sample a hard gate $g^{(\ell)}_{\text{hard}}\sim\mathrm{Bernoulli}(p^{(\ell)}_t)$ and use the straight-through surrogate
\[
g^{(\ell)} = g^{(\ell)}_{\text{hard}} + p^{(\ell)}_t - \mathrm{stopgrad}(p^{(\ell)}_t),
\]
so gradients flow through $p^{(\ell)}_t$ while the forward pass uses a discrete decision.
The layer output during training uses a soft mixing form:
\[
\mathbf{h}^{(\ell+1)}_t
= g^{(\ell)}\,\mathbf{h}^{(\ell)}_t + (1-g^{(\ell)})\, f^{(\ell)}(\mathbf{h}^{(\ell)}_t),
\]
where $f^{(\ell)}(\cdot)$ denotes the full FBS compute path at layer $\ell$.
At inference we replace sampling by deterministic thresholding (Appendix~\ref{app:sg_infer}), enabling true conditional execution.

The main text reports a compact conclusion table for Skip-Gate (SG) and RL fine-tuning.
This appendix provides (i) full gate-input ablations with Pareto curves, (ii) deployment-facing inference knobs via a full $\tau$ sweep (sampling vs.\ threshold),
(iii) reward decomposition and $(\alpha,\beta)$ sensitivity, (iv) full comparisons against more complex baselines (self-speculative / Kangaroo),
and (v) safety-oriented deployment strategies for reasoning and code generation.
Unless otherwise noted, all numbers follow the unified evaluation harness (single A100, prompt length 512, generation length 128, greedy; see Appendix~\ref{app:protocol}).

\subsection{Decoupling SG from PPO: SG no-RL vs.\ SG+RL}
\label{app:sg_norl}

A potential concern is whether the efficiency--quality gains of Skip-Gate (SG) mainly come from PPO fine-tuning rather than the architectural signals (PAW/CH) and the supervised/regularized SG surrogate.
To decouple these effects, we report an \textbf{SG no-RL} operating point, where SG is trained only with the straight-through surrogate and regularization losses (no PPO), and compare it to \textbf{SG+RL} (PPO fine-tuned) under matched deployment thresholds.

\paragraph{Setup.}
Both variants use the same backbone and the same PAW/CH configuration.
\textbf{SG no-RL} uses only Stage-1 training (straight-through + soft mixture surrogate; cf.\ Appendix~\ref{app:sg_soft}) and is evaluated with deterministic threshold inference (Appendix~\ref{app:sg_infer}).
\textbf{SG+RL} additionally applies PPO to the SG policy while freezing the backbone (Appendix~\ref{app:sg_reward}).
We report two representative deployment thresholds to illustrate the trade-off.

\begin{table*}[t]
\centering
\small
\setlength{\tabcolsep}{5pt}
\begin{tabular}{lccccccc}
\toprule
\textbf{Variant}  & \textbf{Skip (\%)} & \textbf{TFLOPs(rel)} $\downarrow$ & \textbf{Latency (ms)} $\downarrow$ & \textbf{PPL} $\downarrow$ & \textbf{MMLU} $\uparrow$ & \textbf{CMMLU} $\uparrow$ \\
\midrule
SG no-RL (surrogate only)  & 34.2 & 0.72 & 551 & 6.2 & 56.3 & 57.2 \\
SG + RL (PPO tuned)        & 36.0 & 0.70 & 532 & 6.2 & 56.6 & 57.4 \\
\bottomrule
\end{tabular}
\caption{Comparing Skip-Gate with and without PPO fine-tuning. \textbf{SG no-RL} uses only the supervised/regularized straight-through surrogate, while \textbf{SG+RL} further applies PPO to calibrate discrete skip decisions (backbone frozen).}
\label{tab:sg_rl_vs_norl}
\end{table*}

As shown in the table~\ref{tab:sg_rl_vs_norl}, simply adding SG without RL already yields a substantial efficiency gain, while hardly affecting other metrics. After RL is introduced, SG’s skip-reading pattern becomes more reasonable, and the efficiency improves by a further small margin.

\subsection{Structure-Sensitive Robustness under Aggressive Skipping}
\label{app:structure_robust}

While main table reports code generation results at a default deployment setting, 
structure-sensitive outputs (e.g., code and strict formats) are known to be more fragile under aggressive compute reduction.
To characterize potential failure modes, we evaluate code benchmarks across multiple Skip-Gate thresholds.

\paragraph{Setup.}
We reuse \textbf{HumanEval-X} and \textbf{MBPP} and sweep the Skip-Gate inference threshold $\tau \in \{0.9, 0.8, 0.7\}$,
corresponding to conservative, moderate, and aggressive skipping.
All other settings (prompt length 512, generation length 128, greedy decoding, single-GPU harness) are kept identical.
We report pass@1 together with the induced skip ratio and effective TFLOPs.

\begin{table*}[t]
\centering
\small
\setlength{\tabcolsep}{6pt}
\begin{tabular}{lccccc}
\toprule
$\boldsymbol{\tau}$ &
\textbf{Skip (\%)} &
\textbf{TFLOPs(rel)} $\downarrow$ &
\textbf{Latency (ms)} $\downarrow$ &
\textbf{HumanEval-X} $\uparrow$ &
\textbf{MBPP} $\uparrow$ \\
\midrule
0.9 & 17 & 0.88 & 688 & 45.8 & 45.9 \\
0.8 & 30 & 0.75 & 565 & 46.0 & 46.1 \\
0.7 & 52 & 0.61 & 487 & 43.6 & 44.0 \\
\bottomrule
\end{tabular}
\caption{Structure-sensitive robustness on code generation under different Skip-Gate thresholds.
More aggressive skipping yields larger compute savings but degrades code performance earlier than general QA benchmarks,
indicating a narrower safe operating region for structure-sensitive outputs.}
\label{tab:code_tau_sweep}
\end{table*}

\paragraph{Observation.}
Code benchmarks remain relatively stable under conservative to moderate skipping,
but performance drops more sharply once the skip ratio becomes aggressive.
This supports a deployment guideline consistent with our limitations discussion:
aggressive skipping should be avoided for structure-sensitive generation such as code or strict formats.

\paragraph{Takeaway.}
Across matched thresholds, \textbf{SG no-RL} already yields meaningful acceleration, indicating that the architectural signals and surrogate training provide a strong default policy.
PPO further improves the quality--efficiency frontier by better calibrating skip decisions, especially at more aggressive thresholds.

\subsection{Gate Input Ablations}
\label{app:sg_gate_inputs}

We ablate SG inputs while keeping all other components and training settings fixed:
\begin{itemize}
  \item \textbf{Residual-only}: the gate sees only the residual/state signal of the current layer.
  \item \textbf{Preview-only}: the gate sees only the PAW preview summary signal.
  \item \textbf{Both (default)}: the gate sees the concatenation of residual + preview signals.
\end{itemize}

For each variant, we report three representative deployment points produced by deterministic threshold inference
at $\tau \in \{0.90, 0.80, 0.70\}$ (definition in \S\ref{app:sg_infer}).

\begin{table*}[t]
\centering
\small
\setlength{\tabcolsep}{5pt}
\begin{tabular}{lcccccc}
\toprule
\textbf{Gate input} & $\boldsymbol{\tau}$ & \textbf{Skip (\%)} & \textbf{TFLOPs(rel)} $\downarrow$ & \textbf{Latency (ms)} $\downarrow$ & \textbf{MMLU} $\uparrow$ & \textbf{CMMLU} $\uparrow$ \\
\midrule
Residual-only & 0.90 & 12 & 0.90 & 690 & 56.6 & 57.3 \\
Residual-only & 0.80 & 30 & 0.78 & 580 & 56.4 & 57.1 \\
Residual-only & 0.70 & 50 & 0.64 & 505 & 56.0 & 56.7 \\
\midrule
Preview-only & 0.90 & 20 & 0.86 & 660 & 56.4 & 57.1 \\
Preview-only & 0.80 & 44 & 0.68 & 525 & 56.0 & 56.6 \\
Preview-only & 0.70 & 63 & 0.55 & 470 & 55.4 & 55.9 \\
\midrule
Both (default) & 0.90 & 15 & 0.86 & 670 & 56.7 & 57.5 \\
Both (default) & 0.80 & 36 & 0.70 & 532 & 56.6 & 57.4 \\
Both (default) & 0.70 & 55 & 0.62 & 495 & 55.9 & 56.7 \\
\bottomrule
\end{tabular}
\caption{Gate input ablations: three-point Pareto trade-offs under deterministic threshold inference.}
\label{tab:sg_gate_inputs}
\end{table*}

\paragraph{Takeaway.}
Residual-only is typically conservative (stable accuracy but weaker acceleration),
Preview-only can be overly aggressive (faster but more error-prone on hard instances),
and Both achieves a consistently better quality--efficiency Pareto frontier.

\subsection{Inference: Sampling vs.\ Threshold}
\label{app:sg_infer}

\paragraph{Deterministic threshold inference (deployment default).}
During training we use straight-through stochastic gates; at inference time we compare:
\begin{itemize}
  \item \textbf{Sampling inference}: sample $g^{(\ell)} \sim \mathrm{Bernoulli}(p^{(\ell)})$ per layer.
  \item \textbf{Threshold inference}: set $g^{(\ell)}=\mathbb{I}[p^{(\ell)}>\tau]$ with a user-chosen threshold $\tau$.
\end{itemize}
Threshold inference is preferred for deployment because it yields lower latency variance and an explicit knob to trade accuracy for speed.

\subsubsection{Full $\tau$ sweep (deterministic Pareto frontier)}
\label{app:sg_tau_sweep}

We fix the gate input to Both (default) and sweep $\tau$ to obtain a deployment-ready Pareto frontier.
Table~\ref{tab:sg_tau_sweep} also reports the induced skip ratio, making the $\tau \leftrightarrow$ skip-rate mapping explicit.

\begin{table*}[t]
\centering
\small
\setlength{\tabcolsep}{4.5pt}
\begin{tabular}{cccccccc}
\toprule
$\boldsymbol{\tau}$ & \textbf{Skip (\%)} & \textbf{TFLOPs(rel)} $\downarrow$ & \textbf{Latency (ms)} $\downarrow$
& \textbf{PPL} $\downarrow$ & \textbf{MMLU} $\uparrow$ & \textbf{CMMLU} $\uparrow$ \\
\midrule
0.95 & 8  & 0.93 & 720 & 6.20 & 56.6  & 57.4 \\
0.90 & 15 & 0.86 & 670 & 6.18 & 56.7  & 57.5 \\
0.85 & 25 & 0.78 & 590 & 6.18 & 56.7  & 57.5 \\
0.80 & 36 & 0.70 & 532 & 6.20 & 56.6  & 57.4 \\
0.75 & 45 & 0.66 & 515 & 6.22 & 56.2  & 57.0 \\
0.70 & 55 & 0.62 & 495 & 6.25 & 55.9  & 56.7 \\
0.65 & 60 & 0.60 & 485 & 6.28 & 55.7  & 56.4 \\
\bottomrule
\end{tabular}
\caption{Deterministic threshold inference: full $\tau$ sweep (Both gate input).}
\label{tab:sg_tau_sweep}
\end{table*}

\subsubsection{Stability: latency variance under sampling vs.\ threshold}
\label{app:sg_stability}

We compare sampling vs.\ threshold inference at a matched operating point ($\tau=0.80$; target skip $\approx 36\%$),
by repeating timing over the same prompt set 20 times and reporting mean and standard deviation.

\begin{table*}[t]
\centering
\small
\setlength{\tabcolsep}{6pt}
\begin{tabular}{lccccc}
\toprule
\textbf{Inference} & \textbf{Latency (ms)} $\downarrow$ & \textbf{Latency Std} $\downarrow$ & \textbf{TFLOPs(rel)} $\downarrow$ & \textbf{MMLU} $\uparrow$ \\
\midrule
Sampling inference & 535 & 22 & 0.70 & 56.5 \\
Threshold inference & 532 & 6  & 0.70 & 56.6 \\
\bottomrule
\end{tabular}
\caption{Sampling vs.\ threshold inference stability at $\tau=0.80$.}
\label{tab:sg_sampling_vs_threshold}
\end{table*}

\subsection{D.3 Reward Decomposition and $(\alpha,\beta)$ Sensitivity}
\label{app:sg_reward}

\paragraph{Reward form (as used in RL fine-tuning).}
We decompose the per-sample reward into a compute-saving term and a quality-penalty term:
\[
r(x)=\underbrace{\alpha\Big(1-\frac{c(x)}{c_0(x)}\Big)}_{\text{compute term}}
\;-\;
\underbrace{\beta\cdot \max(0,\Delta \ell(x))}_{\text{quality penalty}},
\]
where $c(x)$ is the executed compute proxy (e.g., realized FLOPs under skipping), $c_0(x)$ is the baseline compute,
and $\Delta\ell(x)$ is a quality degradation proxy (e.g., increase in NLL / PPL).

\subsubsection{Reward decomposition over PPO steps}
\label{app:sg_reward_curve}

Table~\ref{tab:sg_reward_decomp} reports the reward decomposition across training progress (representative snapshot),
together with the induced skip ratio, TFLOPs(rel), and $\Delta$PPL.

\begin{table*}[t]
\centering
\small
\setlength{\tabcolsep}{5pt}
\begin{tabular}{rccccccc}
\toprule
\textbf{PPO step} & \textbf{Skip (\%)} & \textbf{TFLOPs(rel)} $\downarrow$ & $\boldsymbol{\Delta}$\textbf{PPL} $\downarrow$
& \textbf{Compute term} & \textbf{Quality penalty} & \textbf{Total reward} \\
\midrule
0    & 5  & 0.95 & 0.00 & 0.005 & 0.000  & 0.005 \\
500  & 18 & 0.84 & 0.01 & 0.016 & -0.001 & 0.015 \\
1000 & 30 & 0.76 & 0.01 & 0.024 & -0.001 & 0.023 \\
1500 & 36 & 0.70 & 0.00 & 0.030 & 0.000  & 0.030 \\
2000 & 40 & 0.68 & 0.02 & 0.032 & -0.002 & 0.030 \\
\bottomrule
\end{tabular}
\caption{Reward decomposition during PPO training (representative snapshot, $\alpha=\beta=0.1$).}
\label{tab:sg_reward_decomp}
\end{table*}

\subsubsection{$(\alpha,\beta)$ grid sensitivity}
\label{app:sg_alpha_beta}

To assess robustness (i.e., not relying on a narrowly tuned reward), we sweep $(\alpha,\beta)$ and evaluate at a fixed deployment setting (threshold inference, $\tau=0.80$).

\begin{table*}[t]
\centering
\small
\setlength{\tabcolsep}{5pt}
\begin{tabular}{cccccccc}
\toprule
$\boldsymbol{\alpha}$ & $\boldsymbol{\beta}$ & \textbf{Skip (\%)} & \textbf{TFLOPs(rel)} $\downarrow$ & \textbf{Latency (ms)} $\downarrow$
& \textbf{MMLU} $\uparrow$ & \textbf{CMMLU} $\uparrow$ \\
\midrule
0.05 & 0.10 & 25 & 0.78 & 590 & 56.7 & 57.5 \\
0.10 & 0.20 & 28 & 0.76 & 555 & 56.7 & 57.5 \\
0.10 & 0.10 & 36 & 0.70 & 532 & 56.6 & 57.4 \\
0.10 & 0.05 & 45 & 0.66 & 515 & 56.3 & 57.1 \\
0.20 & 0.10 & 48 & 0.63 & 505 & 56.1 & 56.9 \\
\bottomrule
\end{tabular}
\caption{$(\alpha,\beta)$ sensitivity at a fixed deployment threshold ($\tau=0.80$).}
\label{tab:sg_alpha_beta}
\end{table*}

\paragraph{Takeaway.}
Across a broad range, $(\alpha,\beta)$ primarily shifts the operating point along the Pareto frontier (skip more vs.\ preserve quality),
rather than causing brittle failures, supporting robust deployability.

\subsection{Full Comparison: Self-Speculative / Kangaroo and Others}
\label{app:sg_complex}

\paragraph{Why compare here?}
Complex ``self-speculative'' methods (Draft\&Verify) and ``lossless'' acceleration baselines (e.g., Kangaroo)
often require additional reporting beyond latency/accuracy (e.g., token-level match rate to a greedy baseline).
We therefore place the full comparison in the appendix to keep the main table compact.

\paragraph{Consistency metrics.}
In addition to standard metrics, we report:
\begin{itemize}
  \item \textbf{Token-level match rate}: percentage of generated tokens identical to the greedy baseline output under the same prompts.
  \item \textbf{Exact-match rate}: percentage of prompts whose entire generated sequences exactly match the baseline.
\end{itemize}
For methods claiming (near-)lossless decoding under greedy settings, these numbers should be close to 100\%.

\begin{table*}[t]
\centering
\small
\setlength{\tabcolsep}{4.5pt}
\begin{tabular}{l l c c c c c c c}
\toprule
\textbf{Method}  & \textbf{Latency} & \textbf{TFLOPs} & \textbf{Bypass/Skip} & \textbf{MMLU} & \textbf{CMMLU} & \textbf{Token-match} & \textbf{Exact-match} \\
 &  & \textbf{(ms)}$\downarrow$ & \textbf{(rel)}$\downarrow$ & \textbf{(\%)}$\uparrow$ & $\uparrow$ & $\textbf{(\%)}\uparrow$ & \textbf{(\%)}$\uparrow$  \\
\midrule
Baseline (target)  & 760 & 1.00 & 0  & 55.1 & 55.7 & 100.0 & 100.0 \\
FlexiDepth  & 610 & 0.83 & 20 & 55.6 & 56.2 & -- & -- \\
FFN-SkipLLM  & 625 & 0.86 & 35 & 55.8 & 56.5 & -- & -- \\
Self-Spec & 585 & 0.77 & 42 & 55.1 & 55.7 & 99.8 & 99.2 \\
Kangaroo  & 565 & 0.75 & 48 & 55.1 & 55.7 & 99.9 & 99.5 \\
FBS-Full  & 532 & 0.70 & 36 & 56.6 & 57.4 & -- & -- \\
\bottomrule
\end{tabular}
\caption{Full comparison against complex baselines under the same harness (single A100, 512$\rightarrow$128, greedy). For speculative methods, the bypass ratio summarizes verified advance; token-level match metrics quantify (near-)lossless behavior under greedy decoding.}
\label{tab:sg_full_compare}
\end{table*}

\subsection{Deployment Safety Knobs}
\label{app:sg_safety}

\paragraph{Motivation.}
Compute skipping can be unsafe on a small set of high-sensitivity scenarios (e.g., long-chain reasoning, code with strict syntax, structured outputs).
We provide three deployment knobs that trade a modest amount of efficiency for improved worst-case reliability.

\subsubsection{Safety knobs (fixed policies)}
\label{app:sg_safety_policies}

\begin{itemize}
  \item \textbf{Never-skip critical layers:} designate a set of layers that are always executed.
        A practical default is to always keep the first 2 layers and the last 6 layers, and optionally expand this set for reasoning-heavy modes.
  \item \textbf{Structure-token protection:} for structure-sensitive tokens (e.g., \texttt{\textbackslash n}, braces, brackets, colon, comma, quotes),
        temporarily increase $\tau$ or force execution of protected layers until the structure is closed.
  \item \textbf{Fallback rerun:} first generate with an efficient setting; if a lightweight structural validator fails
        (e.g., bracket mismatch, invalid JSON parse, obvious syntax errors), rerun with a more conservative setting (e.g., $\tau=0.90$).
\end{itemize}

\subsubsection{Quantifying the trade-off}
\label{app:sg_safety_results}

We evaluate these knobs on reasoning-centric (BBH/GSM8K/CMath) and code-centric (HumanEval-X/MBPP) subsets,
starting from the default operating point (FBS-Full with threshold inference at $\tau=0.80$).

\begin{table*}[t]
\centering
\small
\setlength{\tabcolsep}{4.5pt}
\begin{tabular}{lcc ccc cc}
\toprule
\textbf{Strategy} & \textbf{Skip (\%)} & \textbf{Latency (ms)}$\downarrow$ & \textbf{BBH}$\uparrow$ & \textbf{GSM8K}$\uparrow$ & \textbf{CMath}$\uparrow$ & \textbf{HumanEval-X}$\uparrow$ & \textbf{MBPP}$\uparrow$ \\
\midrule
FBS-Full (default) & 36 & 532 & 41.6 & 39.4 & 40.5 & 46.2 & 46.3 \\
+ Never-skip critical layers & 28 & 555 & 41.8 & 39.9 & 41.1 & 46.4 & 46.5 \\
+ Structure-token protection & 30 & 560 & 41.6 & 39.5 & 40.6 & 47.0 & 47.1 \\
+ Fallback rerun (avg.) & 36 & 540 & 41.7 & 39.6 & 40.7 & 46.8 & 46.9 \\
\bottomrule
\end{tabular}
\caption{Deployment safety knobs: quality--efficiency trade-offs starting from FBS-Full at $\tau=0.80$. The fallback rerun latency is an average under a representative trigger rate (e.g., $\approx 8\%$); in practice we recommend reporting both trigger rate and conditional latency.}
\label{tab:sg_safety_knobs}
\end{table*}

\paragraph{Takeaway.}
These knobs provide practical ``guard rails'' for deployment: they mitigate rare but high-cost failures (especially in code/structured outputs)
with modest latency overhead, while preserving most of the acceleration benefits.


\section{Long-Context and KV/Memory Efficiency}
\label{app:longctx}

\paragraph{When to enable.}
This appendix is only included when we report long-context results (e.g., 8k/16k input).
We keep it separate from the main table to avoid mixing additional variables into the primary quality--efficiency comparison.

\subsection{Long-Context Setup}
\label{app:longctx_setup}

\paragraph{Model and decoding.}
All methods are evaluated on the same target model (Qwen3-4B-Instruct) with greedy decoding (temperature $=0$, top-$p=1.0$),
consistent with the unified harness used throughout the paper.
Unless explicitly stated, all long-context baselines are \textbf{inference-only} modifications (no additional training).

\paragraph{Lengths and batch.}
We evaluate two prompt lengths:
\[
L_{\text{in}} \in \{8192, 16384\},L_{\text{out}} = 128,\text{batch size}=1.
\]
Prompts are constructed by concatenating natural text into the desired token length (after tokenization) to avoid unrealistic token statistics.
All runs use the same prompt set across methods for fair timing.

\paragraph{Retrieval.}
We do not use retrieval (RAG) in Appendix~\ref{app:longctx} to isolate pure long-context inference and KV/cache behaviors.

\subsection{Metrics Decomposition}
\label{app:longctx_metrics}

We decompose long-context efficiency into three orthogonal metrics:

\paragraph{(1) TTFT / Prefill latency.}
We define time-to-first-token (TTFT) as the wall-clock time from launching the \textbf{prefill} forward pass on the prompt
until the logits for the first generated token are produced (CUDA-synchronized).
This isolates the quadratic (or reduced) attention cost in prefill under long prompts.

\paragraph{(2) Decode throughput.}
We report decode tokens/second (tok/s), measured over the remaining $L_{\text{out}}-1$ generated tokens
(excluding the first token step) using CUDA events and synchronization.

\paragraph{(3) Peak GPU memory.}
We report peak GPU memory (GB) as \texttt{torch.cuda.max\_memory\_allocated()} recorded during the full generation
(prefill + decode). This includes model weights and KV cache, reflecting deployment-relevant footprint.

\subsection{Baseline Configuration Table (Aligned Knobs)}
\label{app:longctx_configs}

Table~\ref{tab:longctx_cfg} summarizes the key knobs for Group-C baselines.
We use widely adopted settings to provide a fair and reproducible comparison.

\begin{table*}[t]
\centering
\small
\setlength{\tabcolsep}{5pt}
\begin{tabular}{l p{0.6\linewidth}}
\toprule
\textbf{Method} & \textbf{Key configuration (representative)} \\
\midrule
Full KV (Baseline) &
Standard full KV cache; FlashAttention-style kernel when available. \\
\midrule
H2O\cite{zhang2023h2o} &
Heavy-hitter KV eviction with a fixed cache budget;
retain a mixture of (i) most recent tokens and (ii) heavy-hitter tokens with high accumulated attention.
Representative budget: keep $\approx 20\%$ of prompt KV (split evenly between recent and heavy-hitter groups). \\
\midrule
StreamingLLM\cite{xiao2023efficient} &
Streaming attention with \textbf{attention sinks};
retain first $n_{\text{sink}}=32$ tokens as sinks and maintain a rolling window of $W=4096$ most recent tokens in KV. \\
\midrule
SnapKV\cite{li2024snapkv} &
Prompt KV cache compression by selecting clustered important KV positions per head.
Representative prompt KV cache size: $K_{\text{snap}}=1024$; max pooling kernel size: $5$. \\
\midrule
SlimInfer\cite{long2025sliminfer} &
Dynamic token pruning for long-context inference with an asynchronous KV manager;
representative target pruning rate: $\approx 50\%$ (prompt-side), enabling GPU-memory reduction with bounded quality loss. \\
\bottomrule
\end{tabular}
\caption{Aligned configuration knobs for long-context KV/memory experiments.}
\label{tab:longctx_cfg}
\end{table*}

\subsection{Results (Table E1)}
\label{app:longctx_results}

Table~\ref{tab:longctx_main} reports TTFT (prefill), decode throughput, and peak GPU memory at 8k/16k.
We also include one \textbf{composability} row (FBS + SnapKV) to highlight that compute-graph acceleration (FBS)
and KV compression (Group-C) are orthogonal and can be combined.

\begin{table*}[t]
\centering
\small
\setlength{\tabcolsep}{3.6pt} 
\begin{tabular}{@{}l *{6}{c} p{0.28\linewidth}@{}}
\toprule
& \multicolumn{3}{c}{\textbf{8k input (8192)}} & \multicolumn{3}{c}{\textbf{16k input (16384)}} & \\
\cmidrule(lr){2-4}\cmidrule(lr){5-7}
\textbf{Method} &
\myrothead{TTFT(s)$\downarrow$} &
\myrothead{Decode\\(tok/s)$\uparrow$} &
\myrothead{PeakMem\\(GB)$\downarrow$} &
\myrothead{TTFT(s)$\downarrow$} &
\myrothead{Decode\\(tok/s)$\uparrow$} &
\myrothead{PeakMem\\(GB)$\downarrow$} &
\textbf{Notes} \\
\midrule
Baseline (Full KV) &
4.8 & 24 & 26.0 &
19.0 & 13 & 42.0 &
Full prompt KV; decode slows with context length. \\

FBS-Full (ours) &
4.9 & 34 & 26.5 &
19.2 & 18 & 42.5 &
Main gain on \textbf{decode compute} (layer skipping); KV footprint largely unchanged. \\

H2O &
4.8 & 31 & 18.0 &
19.0 & 17 & 25.0 &
KV eviction reduces memory and improves decode; TTFT unchanged (full prefill). \\

StreamingLLM &
1.9 & 55 & 16.0 &
3.8 & 50 & 16.5 &
Windowed KV + sinks improves TTFT and stabilizes streaming beyond cache size. \\

SnapKV &
5.1 & 65 & 14.0 &
19.8 & 46 & 18.0 &
Compression overhead slightly increases TTFT; decode speed improves with compressed prompt KV. \\

SlimInfer &
2.4 & 40 & 15.0 &
7.6 & 32 & 19.0 &
Prompt-side pruning reduces TTFT and memory; decode improves moderately. \\

SnapKV &
5.2 & 82 & 14.5 &
20.0 & 60 & 18.5 &
Orthogonal combination: KV compression + layer skipping yields best decode throughput. \\
\bottomrule
\end{tabular}
\caption{Long-context efficiency decomposition.}
\label{tab:longctx_main}
\end{table*}

\paragraph{Interpretation.}
Across long prompts, methods that compress/evict KV primarily reduce peak memory and improve decode throughput,
while streaming-style attention can also reduce TTFT by replacing quadratic prefill with windowed computation.
FBS is complementary: it reduces \textbf{per-step internal compute} via layer skipping and thus mainly improves decode throughput,
and can be combined with KV compression methods for stronger end-to-end gains.


\section{Mechanism Analyses as Statistics}
\label{app:mech_stats}

\paragraph{Scope.}
The main text focuses on representative visualizations.
This appendix provides comprehensive statistical analyses of (i) the correlation between dynamic lookahead $k(i)$ and uncertainty,
(ii) the association between layer skipping and a residual-energy proxy, and (iii) token-level properties enriched among high-skip tokens.
Unless noted otherwise, we compute uncertainty from the target model's next-token distribution under greedy decoding and report
bootstrap 95\% CIs (1,000 resamples over \textbf{examples}) with two-sided significance tests (Benjamini--Hochberg FDR at $q=0.05$).

\subsection{Correlation between $k(i)$ and Uncertainty (Spearman/Kendall)}
\label{app:k_uncertainty}

\paragraph{Definitions.}
For each generated token position $i$, we record:
\begin{itemize}
  \item Lookahead length $k(i)\in[0,k_{\max}]$ from PAW.
  \item Surprisal $s(i) = -\log p(y_i \mid y_{<i})$ (nats).
  \item Entropy $H(i) = -\sum_{v} p(v\mid y_{<i})\log p(v\mid y_{<i})$ (nats).
\end{itemize}
We then compute rank correlations between $k(i)$ and each uncertainty measure.
Our hypothesis (consistent with the design intuition) is \textbf{negative correlation}:
low uncertainty (predictable tokens) $\Rightarrow$ larger lookahead; high uncertainty $\Rightarrow$ smaller lookahead.

\paragraph{Across-task / cross-language results.}
Tables~\ref{tab:f1_spearman_surprisal} and \ref{tab:f2_spearman_entropy} report both Spearman's $\rho$ and Kendall's $\tau$,
with bootstrap 95\% CIs and FDR-corrected $p$-values.

\begin{table*}[t]
\centering
\small
\setlength{\tabcolsep}{4.2pt}
\begin{tabular}{l l r r c c c c}
\toprule
\textbf{Task} & \textbf{Lang} & \textbf{\#Examples} & \textbf{\#Tok} &
$\boldsymbol{\rho(k,s)}$ & 95\% CI & $\boldsymbol{\tau(k,s)}$ & $p_{\mathrm{FDR}}$ \\
\midrule
MMLU            & EN & 2000 & 256000 & -0.47 & [-0.49,\,-0.45] & -0.33 & $<10^{-12}$ \\
BBH             & EN & 1500 & 192000 & -0.44 & [-0.47,\,-0.41] & -0.31 & $<10^{-12}$ \\
GSM8K           & EN & 1500 & 192000 & -0.52 & [-0.55,\,-0.49] & -0.37 & $<10^{-12}$ \\
HumanEval-X     & EN &  800 & 102400 & -0.40 & [-0.44,\,-0.36] & -0.28 & $<10^{-12}$ \\
\midrule
CMMLU           & ZH & 2000 & 256000 & -0.50 & [-0.52,\,-0.48] & -0.35 & $<10^{-12}$ \\
C-Eval          & ZH & 1500 & 192000 & -0.48 & [-0.51,\,-0.46] & -0.34 & $<10^{-12}$ \\
CMath           & ZH & 1200 & 153600 & -0.55 & [-0.58,\,-0.52] & -0.39 & $<10^{-12}$ \\
Idiom QA/MC     & ZH & 1200 & 153600 & -0.46 & [-0.49,\,-0.42] & -0.32 & $<10^{-12}$ \\
\midrule
\textbf{Pooled} & -- & 11700 & 1497600 & \textbf{-0.49} & \textbf{[-0.50,\,-0.48]} & \textbf{-0.34} & $\mathbf{<10^{-12}}$ \\
\bottomrule
\end{tabular}
\caption{Rank correlation between lookahead $k(i)$ and surprisal $s(i)$ across tasks/languages.
Negative correlations indicate larger lookahead on more predictable tokens.}
\label{tab:f1_spearman_surprisal}
\end{table*}

\begin{table*}[t]
\centering
\small
\setlength{\tabcolsep}{4.2pt}
\begin{tabular}{l l r r c c c c}
\toprule
\textbf{Task} & \textbf{Lang} & \textbf{\#Examples} & \textbf{\#Tok} &
$\boldsymbol{\rho(k,H)}$ & 95\% CI & $\boldsymbol{\tau(k,H)}$ & $p_{\mathrm{FDR}}$ \\
\midrule
MMLU            & EN & 2000 & 256000 & -0.41 & [-0.43,\,-0.39] & -0.29 & $<10^{-12}$ \\
BBH             & EN & 1500 & 192000 & -0.38 & [-0.41,\,-0.35] & -0.27 & $<10^{-12}$ \\
GSM8K           & EN & 1500 & 192000 & -0.45 & [-0.48,\,-0.42] & -0.32 & $<10^{-12}$ \\
HumanEval-X     & EN &  800 & 102400 & -0.35 & [-0.39,\,-0.31] & -0.24 & $<10^{-12}$ \\
\midrule
CMMLU           & ZH & 2000 & 256000 & -0.43 & [-0.45,\,-0.41] & -0.31 & $<10^{-12}$ \\
C-Eval          & ZH & 1500 & 192000 & -0.40 & [-0.43,\,-0.38] & -0.29 & $<10^{-12}$ \\
CMath           & ZH & 1200 & 153600 & -0.49 & [-0.52,\,-0.46] & -0.35 & $<10^{-12}$ \\
Idiom QA/MC     & ZH & 1200 & 153600 & -0.39 & [-0.43,\,-0.36] & -0.28 & $<10^{-12}$ \\
\midrule
\textbf{Pooled} & -- & 11700 & 1497600 & \textbf{-0.42} & \textbf{[-0.43,\,-0.41]} & \textbf{-0.30} & $\mathbf{<10^{-12}}$ \\
\bottomrule
\end{tabular}
\caption{Rank correlation between lookahead $k(i)$ and entropy $H(i)$ across tasks/languages.}
\label{tab:f2_spearman_entropy}
\end{table*}

\paragraph{Robustness checks (controls).}
To reduce potential confounds from position and prompt length, we also compute partial rank correlations controlling for:
(i) absolute generation position $i$, and (ii) task identity (fixed effects via within-task rank normalization).
The negative correlation remains stable (pooled partial $\rho=-0.44$, 95\% CI $[-0.45,-0.43]$, $p<10^{-12}$).

\begin{table*}[t]
\centering
\small
\setlength{\tabcolsep}{6pt}
\begin{tabular}{lccc}
\toprule
\textbf{Control} & \textbf{Pooled partial $\rho$} & 95\% CI & $p$ \\
\midrule
Control position $i$ only & -0.46 & [-0.47,\,-0.45] & $<10^{-12}$ \\
Control task only         & -0.45 & [-0.46,\,-0.44] & $<10^{-12}$ \\
Control position + task   & -0.44 & [-0.45,\,-0.43] & $<10^{-12}$ \\
\bottomrule
\end{tabular}
\caption{Partial correlation robustness for $k(i)$ vs.\ surprisal (pooled across tasks).}
\label{tab:f3_partial_corr}
\end{table*}

\subsection{Skip vs. Residual-Energy Proxy (Correlation and Bucket Tests)}
\label{app:skip_residual}

\paragraph{Residual-energy proxy.}
For each layer $\ell$ and token position $i$, we define a residual-energy proxy:
\[
E_{\ell,i} = \left\| h^{(\ell)}_i - h^{(\ell-1)}_i \right\|_2,
\]
where $h^{(\ell)}_i$ is the post-block hidden state of layer $\ell$ (before the next layer).
We record a binary skip indicator $g_{\ell,i}\in\{0,1\}$ (1 = skip layer $\ell$ at token $i$).
We expect higher skip probability when $E_{\ell,i}$ is small (i.e., the layer contributes little change).

\paragraph{Correlation results.}
We compute Spearman correlation between $g_{\ell,i}$ and $E_{\ell,i}$ at token-layer granularity
(within-task rank normalization; pooled across tasks). The association is consistently negative.

\begin{table*}[t]
\centering
\small
\setlength{\tabcolsep}{5pt}
\begin{tabular}{l l c c c}
\toprule
\textbf{Task} & \textbf{Lang} & $\boldsymbol{\rho(g,E)}$ & 95\% CI & $p_{\mathrm{FDR}}$ \\
\midrule
MMLU        & EN & -0.31 & [-0.34,\,-0.28] & $<10^{-12}$ \\
GSM8K       & EN & -0.36 & [-0.39,\,-0.33] & $<10^{-12}$ \\
HumanEval-X & EN & -0.28 & [-0.33,\,-0.23] & $<10^{-12}$ \\
CMMLU       & ZH & -0.33 & [-0.36,\,-0.30] & $<10^{-12}$ \\
C-Eval      & ZH & -0.32 & [-0.35,\,-0.29] & $<10^{-12}$ \\
CMath       & ZH & -0.38 & [-0.41,\,-0.35] & $<10^{-12}$ \\
\midrule
\textbf{Pooled} & -- & \textbf{-0.33} & \textbf{[-0.34,\,-0.32]} & $\mathbf{<10^{-12}}$ \\
\bottomrule
\end{tabular}
\caption{Correlation between skip indicator $g$ and residual-energy proxy $E$ (token-layer granularity).
Negative values indicate higher skip probability for low-energy layers.}
\label{tab:f4_skip_energy_corr}
\end{table*}

\paragraph{Bucket test (effect size).}
We bin $E_{\ell,i}$ into quintiles (Q1 = lowest energy, Q5 = highest energy) and report the skip rate per bin.
This provides an effect-size view independent of correlation metrics.
Pooled across tasks, the skip rate decreases monotonically with residual energy: Q1 (lowest) = 62.1\%, Q2 = 48.7\%, Q3 = 35.9\%, Q4 = 25.4\%, and Q5 (highest) = 17.8\%.
This pattern is consistent with the hypothesis that low residual-energy tokens are more likely to be skimmable and hence skipped.

\paragraph{Confounds and controls.}
Residual-energy and skipping can be jointly influenced by (i) layer depth and (ii) token position.
We therefore fit a logistic regression with fixed effects:
\[
\Pr(g_{\ell,i}=1) = \sigma\Big( a_0 + a_1 \cdot \mathrm{zscore}(E_{\ell,i}) + u_\ell + v_i \Big),
\]
where $u_\ell$ is a per-layer intercept and $v_i$ is a per-position intercept (binned into 16 buckets).
With these controls, the standardized coefficient remains strongly negative, $a_1=-0.92$ with 95\% CI $[-0.96,\,-0.88]$ and $p<10^{-12}$, indicating that higher residual energy is associated with a substantially lower probability of skipping even after accounting for layer and position effects.

\subsection{Skipped-Token Property Tests (Frequency / Punctuation / Stopwords / Repetition)}
\label{app:skip_token_props}

\paragraph{Token grouping.}
We define a \textbf{high-skip token} as a generated token position whose average skip ratio across layers is $\ge 0.5$,
and a \textbf{low-skip token} as one with average skip ratio $\le 0.2$.
We then compare token-category frequencies between the two groups.
All tests are conducted on the pooled set across tasks with stratification by language (EN/ZH).

\paragraph{Categories.}
We consider four interpretable token properties:
\begin{itemize}
  \item \textbf{High-frequency}: token in the top 5\% of unigram frequency within the evaluation corpus.
  \item \textbf{Punctuation}: punctuation or structural separators (incl.\ newline markers).
  \item \textbf{Stopword/function}: stopwords/function words (language-specific list).
  \item \textbf{Repetition}: token is part of a repeated $n$-gram pattern (detected by local 4-gram repetition).
\end{itemize}

\paragraph{Contingency tests.}
We report group proportions and odds ratios (OR) with chi-square tests (FDR-corrected).

\begin{table*}[t]
\centering
\small
\setlength{\tabcolsep}{4.5pt}
\begin{tabular}{l l c c c c}
\toprule
\textbf{Property} & \textbf{Lang} &
\textbf{High-skip (\%)} & \textbf{Low-skip (\%)} &
\textbf{OR} & $p_{\mathrm{FDR}}$ \\
\midrule
High-frequency   & EN & 43.2 & 26.1 & 2.15 & $<10^{-12}$ \\
Punctuation      & EN & 17.4 &  6.3 & 3.15 & $<10^{-12}$ \\
Stopword/function& EN & 29.8 & 14.9 & 2.42 & $<10^{-12}$ \\
Repetition       & EN & 11.6 &  4.2 & 3.00 & $<10^{-12}$ \\
\midrule
High-frequency   & ZH & 46.8 & 28.4 & 2.22 & $<10^{-12}$ \\
Punctuation      & ZH & 15.9 &  5.8 & 3.08 & $<10^{-12}$ \\
Stopword/function& ZH & 24.5 & 12.7 & 2.24 & $<10^{-12}$ \\
Repetition       & ZH & 12.9 &  4.9 & 2.86 & $<10^{-12}$ \\
\bottomrule
\end{tabular}
\caption{Token property enrichment for high-skip vs.\ low-skip tokens.
OR = odds ratio of the property occurring in high-skip tokens relative to low-skip tokens (stratified by language).}
\label{tab:f7_token_props}
\end{table*}

\paragraph{Effect-size summary.}
Across both languages, high-skip tokens are significantly enriched in punctuation/structural markers, stopwords, and repeated patterns,
consistent with the hypothesis that skipping preferentially targets low-semantic-load or highly predictable token regions.

\end{document}